\documentclass{article}

\PassOptionsToPackage{numbers, compress}{natbib}

\usepackage[final]{neurips_2021}

\usepackage[utf8]{inputenc} 
\usepackage[T1]{fontenc}    
\usepackage{hyperref}       
\usepackage{url}            
\usepackage{booktabs}       
\usepackage{amsfonts}       
\usepackage{nicefrac}       
\usepackage{microtype}      
\usepackage{xcolor}         

\usepackage{graphicx}
\usepackage{amsmath}
\usepackage{bm}
\usepackage{makecell}
\usepackage{tabularx}
\newcolumntype{Y}{>{\centering\arraybackslash}X}

\newcommand{\etc}{\textit{etc}}

\newcommand{\ie}{\textit{i.e.}}
\newcommand{\eg}{\textit{e.g.}}

\title{Deceive D: Adaptive Pseudo Augmentation for\\GAN Training with Limited Data}

\author{Liming Jiang$^{1}$ \hspace{12pt} Bo Dai$^{1}$ \hspace{12pt} Wayne Wu$^{2}$ \hspace{12pt} Chen Change Loy$^{1}$\thanks{Corresponding author.}\\[2pt]
$^1$S-Lab, Nanyang Technological University \hspace{12pt} $^2$SenseTime Research\\[1pt]
{\tt\small \{liming002, bo.dai, ccloy\}@ntu.edu.sg} \hspace{12pt}
{\tt\small wuwenyan@sensetime.com}
}

\begin{document}

\maketitle


\begin{abstract}
\label{sec:abstract}

Generative adversarial networks (GANs) typically require ample data for training in order to synthesize high-fidelity images.
Recent studies have shown that training GANs with limited data remains formidable due to discriminator overfitting, the underlying cause that impedes the generator's convergence.
This paper introduces a novel strategy called Adaptive Pseudo Augmentation (APA) to encourage healthy competition between the generator and the discriminator.
As an alternative method to existing approaches that rely on standard data augmentations or model regularization,
APA alleviates overfitting by employing the generator itself to augment the real data distribution with generated images, which deceives the discriminator adaptively.
Extensive experiments demonstrate the effectiveness of APA in improving synthesis quality in the low-data regime. 
We provide a theoretical analysis to examine the convergence and rationality of our new training strategy.
APA is simple and effective. It can be added seamlessly to powerful contemporary GANs, such as StyleGAN2, with negligible computational cost. Code: \url{https://github.com/EndlessSora/DeceiveD}.

\end{abstract}


\section{Introduction}
\label{sec:introduction}

\begin{figure}
	\centering
	\includegraphics[width=\linewidth]{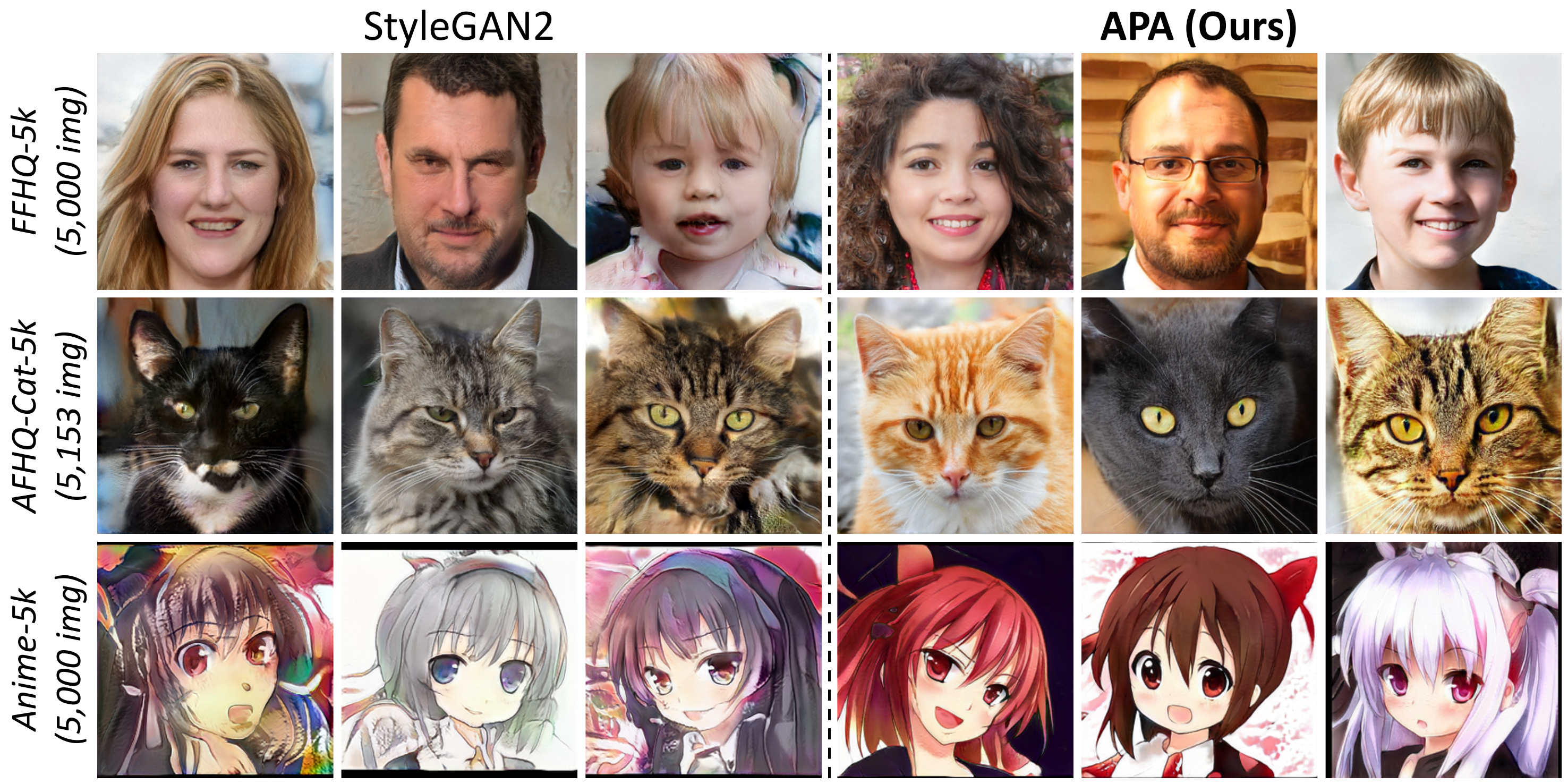}
	\caption{\textbf{StyleGAN2}~\cite{stylegan2} synthesized results (no truncation) deteriorate \textbf{given the limited amount of training data} ($256 \times 256$), \ie, FFHQ~\cite{stylegan}~(a subset of $5,000$ images, $\sim7\%$ of full data), AFHQ-Cat~\cite{starganv2}~($5,153$ images, which is small by itself), and Danbooru2019 Portraits (Anime)~\cite{danbooru2019portraits}~(a subset of $5,000$ images, $\sim2\%$ of full data). The proposed Adaptive Pseudo Augmentation (APA) effectively ameliorates the degraded performance of StyleGAN2 on limited data.}
	\label{fig:teaser}
	\vspace{-0.4cm}
\end{figure}

While state-of-the-art GANs like StyleGAN2~\cite{stylegan2} are constantly pushing forward the fidelity and resolution of synthesized images,
they usually require a large amount of training data to fully unleash their power. Training GANs with insufficient data tend to generate poor-quality images, as shown in Figure~\ref{fig:teaser}.
Collecting sufficient data samples for these GANs is sometimes infeasible,
especially in domains where data are sparse and privacy-sensitive.
To ease the practical deployment of powerful GANs, it is necessary to devise new strategies for training GANs with limited data while preserving the quality of synthesis.
%

Recent studies have shown that the overfitting of the discriminator is the critical reason that impedes effective GAN training on limited data~\cite{noisestable1,ada,da,reglimited},
rendering severe instability of training dynamics.
Specifically,
when the discriminator starts to overfit,
the distributions of its outputs for real and generated samples gradually diverge from each other~\cite{ada,reglimited},
and its feedback to the generator becomes less informative.
Consequently, the generator converges to an inferior point, compromising the quality of synthesized images.
Recent solutions to this problem include the use of standard data augmentations, either conventional or differentiable, to real and generated images~\cite{ada,da,imgaug,ondataauggan} or applying an additional model regularization term~\cite{reglimited}. 
Addressing the discriminator overfitting is still an open problem. We are interested in finding an alternative way to the aforementioned approaches.
%

In this paper, we present a simple yet effective way to regularize a discriminator without introducing any external augmentations or regularization terms.
We call our method \textit{Adaptive Pseudo Augmentation} (APA).
In contrast to previous standard data augmentations~\cite{ada,da,imgaug,ondataauggan}, we exploit the generator in a GAN itself to provide the augmentation, a more natural way to regularize the overfitting of the discriminator.
Compared to the model regularization, our approach is more adaptive to fit different settings and training status without manual tuning.
Specifically, APA takes the fake/pseudo samples synthesized by the generator and moderately feeds them into the limited real data. Such pseudo data are adaptively presented to the discriminator as ``real'' instances.
The goal of this pseudo augmentation for the real data is not to enlarge the real dataset but to suppress the discriminator's confidence in distinguishing real and fake distributions.
The deceit is introduced adaptively, which is moderated by a deception probability according to the degree of overfitting.
To quantify overfitting, we study a series of plausible heuristics derived from the discriminator raw output logits.
%

The main \textbf{contribution} of this work is a novel adaptive pseudo augmentation method for training GANs with limited data. This approach deceives the discriminator adaptively and mitigates the problem of discriminator overfitting.
The proposed APA can be readily added to existing GAN training with negligible computational cost.
We conduct extensive experiments to demonstrate the effectiveness of APA for state-of-the-art GAN training with limited data. The results are comparable or even better than other types of solutions~\cite{ada,reglimited}. APA is also complementary to existing methods based on standard data augmentations for gaining a further performance boost.
Besides, we theoretically connect APA with minimizing the JS divergence~\cite{jsdiv} between the smoothed data distribution and generated distribution, proving its convergence and rationality.
We hope that our approach could extend the breadth and potential of solutions to GAN training with limited data.
%


\section{Related Work}
\label{sec:relatedwork}

\textbf{Generative adversarial networks.}
Generative adversarial networks (GANs)~\cite{GAN,congan,DCGAN,realnessGAN} adopt an adversarial training scheme, where a generator keeps refining its capability in synthesizing images to compete with a discriminator (\ie, a binary classifier) until the discriminator fails to classify the generated samples as fakes.
GANs are known to suffer from training instability~\cite{GAN,improvedtechgans,noisestable1,r1reg}. Various approaches have been proposed to stabilize the training and improve the quality of synthesis by minimizing different $f$-divergences of the real and fake distributions~\cite{fgan}.
The saturated form of vanilla GAN~\cite{GAN} is theoretically proven to minimize the JS divergence~\cite{jsdiv} between the two distributions. LSGAN~\cite{lsgan} and EBGAN~\cite{ebgan} correspond to the optimizations of $\chi^2$-divergence~\cite{x2div} and the total variation~\cite{wgan}, respectively. On another note, WGAN~\cite{wgan} is designed for minimizing the Wasserstein distance.

State-of-the-art methods, such as PGGAN~\cite{pggan}, BigGAN~\cite{BigGAN}, StyleGAN~\cite{stylegan}, and StyleGAN2~\cite{stylegan2}, employ large-scale training with contemporary techniques, achieving photorealistic results.
These methods have been extended to various tasks, including face generation~\cite{pggan,stylegan,stylegan2}, image editing~\cite{faceswap-GAN,stargan,styleclip}, semantic image synthesis~\cite{pix2pixhd,SPADE,CC-FPSE}, image-to-image translation~\cite{pix2pix,cyclegan,starganv2,tsit,ffl,pSp}, style transfer~\cite{UNIT,MUNIT,DRIT}, and GAN inversion~\cite{DGP,pSp,ganinvsurvey}.
Despite the remarkable success, the performance of GANs relies heavily on the amount of training data.
%

\textbf{Training GANs with limited data.}
The significance and difficulty of training GANs with limited data have been attracting attention from many researchers recently.
The issue of data insufficiency tends to cause overfitting in the discriminator ~\cite{detganoverfit,ada,da}, which in turn deteriorates the stability of training dynamics in GANs, compromising the quality of generated images.

Many recent studies~\cite{consisreg,improvedconsisreg,ondataauggan,imgaug,da,ada} propose to apply standard data augmentations for GAN training to enrich the diversity of the dataset to mitigate the overfitting of the discriminator.
For instance, DiffAugment (DA)~\cite{da} adopts the same differentiable augmentation to both real and fake images for the generator and the discriminator without manipulating the target distribution.
Adaptive discriminator augmentation (ADA)~\cite{ada} shares a similar idea with DA, while it further devises an adaptive approach that controls the strength of data augmentations adaptively. In this work, we extend the study of such an adaptive approach.

Another type of solution is model regularization. Previous efforts on regularizing GANs include adding noise to the inputs of the discriminator~\cite{noisestable1,noisestable2,noisestable3}, gradient penalty~\cite{r1reg,wgangp,stablegp}, one-sided label smoothing~\cite{improvedtechgans}, spectral normalization~\cite{spectralnorm}, label noise~\cite{nips16ganworkshop}, \etc.
These methods are designed for stabilizing training or preventing mode collapse~\cite{improvedtechgans}. The essence of their goals could be considered similar to our method since training GANs in the low-data regime exhibits similar behaviors as previously observed in early GANs with sufficient data. Additional discussions on these techniques are provided in the \textit{Appendix}.
Under the limited data setting, a very recent study proposes an LC-regularization term~\cite{reglimited} to regulate the discriminator predictions using two exponential moving average variables that track the discriminator outputs throughout training.

Our work explores an alternative solution from a different perspective, which is also complementary to previous approaches based on standard data augmentations.
%


\section{Methodology}
\label{sec:method}

In GAN's adversarial training, the goal of the generator $G$ is to deceive the discriminator $D$ and maximize the probability that $D$ makes a wrong judgment. Therefore, $G$ keeps refining its generated samples to better deceive $D$ over time.
When the training only accesses a limited amount of data, one would observe that $D$ turns out to be overly confident and hardly makes any mistake, causing its feedback to $G$ to become meaningless.

\begin{figure}
	\centering
	\includegraphics[width=\linewidth]{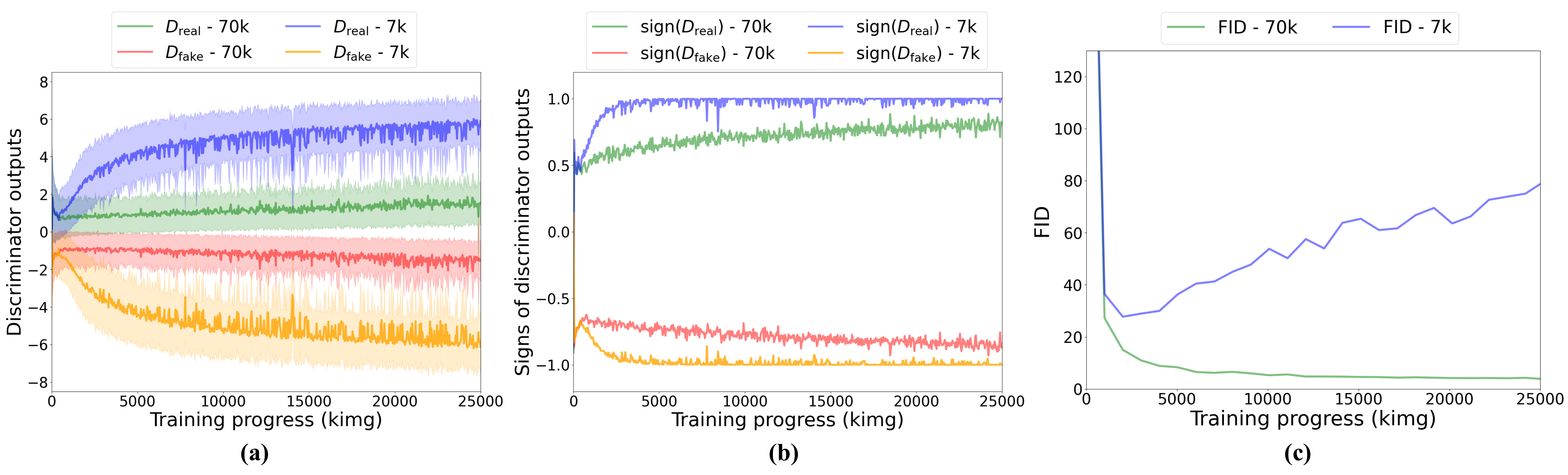}
	\caption{The overfitting of discriminator in GANs when limited training data are available. The three subplots report statistics of training snapshots of two StyleGAN2~\cite{stylegan2} models on FFHQ~\cite{stylegan} ($256 \times 256$). ``70k'' indicates the full dataset, and ``7k'' means a subset of $7,000$ images ($10\%$ data). The ``kimg'' denotes thousands of real images shown to the discriminator. (a) Discriminator raw output logits. (b) Signs of discriminator outputs. (c) Training convergence measured by FID~\cite{TTUR}.}
	\label{fig:doverfit}
	\vspace{-0.3cm}
\end{figure}

In Figure~\ref{fig:doverfit}, we show the training ``snapshots'' of two StyleGAN2~\cite{stylegan2} models on the FFHQ dataset~\cite{stylegan}. The settings of the two models differ only by the amount of data available to them for training.
As can be observed, both training processes start smoothly, and distributions of discriminator outputs for the real and generated images overlap at the early stage. As the training progresses, the discriminator, which only has access to limited data (7k images, $10\%$ of full data), experiences diverged predictions much more rapidly, and the average sign boundary turns out to be more apparent.
This divergence in prediction shows that $D$ becomes increasingly confident in classifying real and fake samples. At the late stage of training, $D$ can even judge all the input samples correctly with high confidence.
Meanwhile, the evaluation FID~\cite{TTUR} scores (lower is better) deteriorate, consistent with the divergence of $D$'s predictions.
The phenomena above demonstrate how a discriminator gets overfitted quickly with limited data. As can be seen from the FID curves, the overfitting directly influences training dynamics and the convergence of $G$, leading to poor generation performance.

\subsection{Adaptive Pseudo Augmentation}
\label{sec:apa}

\begin{figure}[t]
	\centering
	\includegraphics[width=\linewidth]{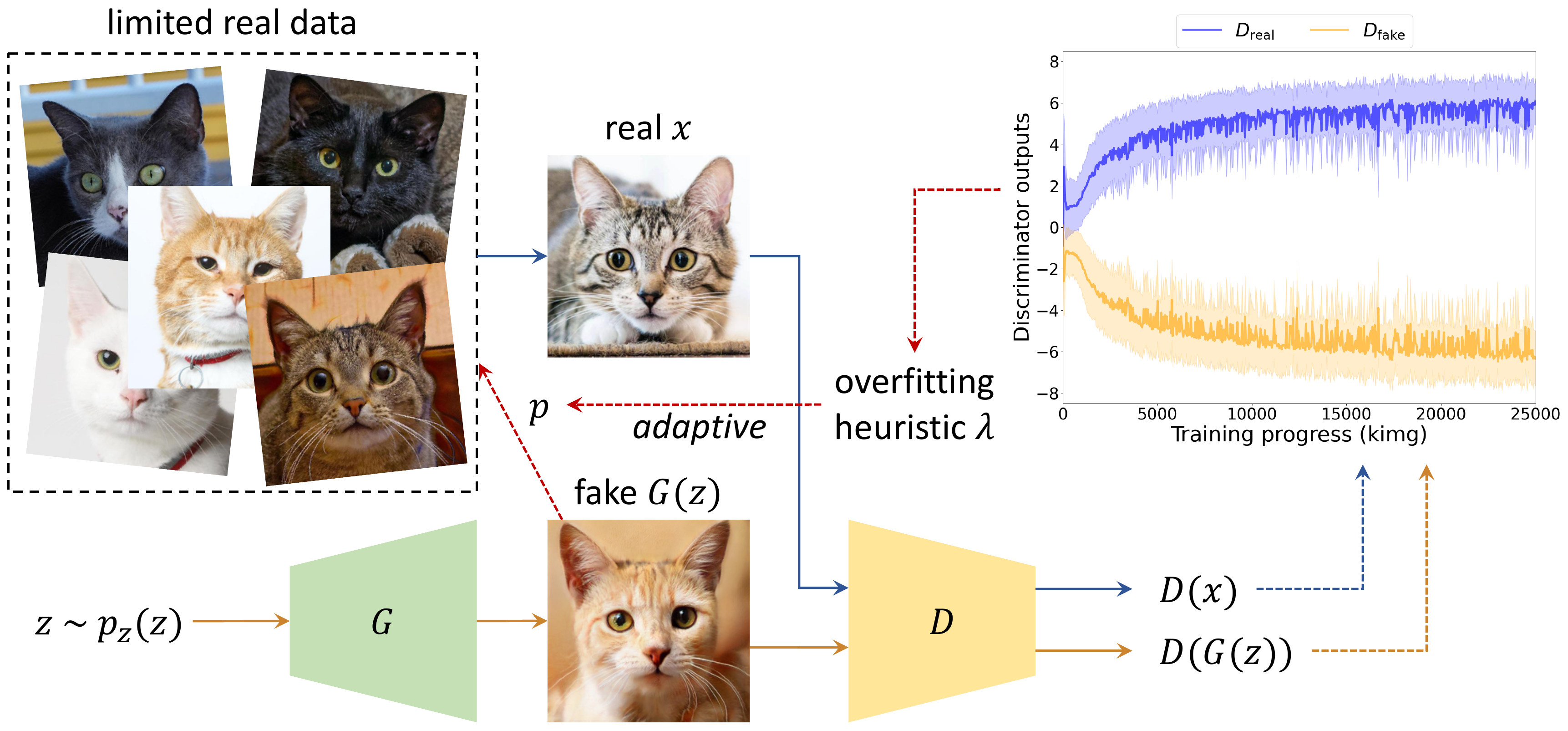}
	\caption{Adaptive pseudo augmentation (APA) for GAN training with limited data. We employ a GAN to augment itself using the generated images to deceive the discriminator adaptively. Specifically, APA feeds the images synthesized by the generator into the limited real data moderately, and these fakes are presented as ``real'' instances to the discriminator. Such deceits are introduced adaptively using an overfitting heuristic $\lambda$ defined by the discriminator raw output logits. The augmentation/deception probability $p$ can be adaptively controlled throughout training.}
	\label{fig:mainidea}
	\vspace{-0.2cm}
\end{figure}

The generator itself naturally possesses the capability to counteract discriminator overfitting.
To harness this capability, our method employs a GAN to augment itself using the generated samples~(see Figure~\ref{fig:mainidea}). Specifically, we feed the samples generated by $G$ into the limited real data to form a pseudo-real set. The fake images in this set will be adaptively presented to $D$, pretending themselves as real data.
The goal here is to deceive $D$ with the pseudo-real set and consequently suppressing its confidence in distinguishing real and fake images.

Blindly presenting the fake images as reals to $D$ may weaken the fundamental ability of $D$ in adversarial training. In our approach, the deceit is introduced adaptively to avoid any potential adverse effects.
To moderate the deception, we perform pseudo augmentation based on a probability, $p\in\left[0,1\right)$, that quantifies the deception strength. Specifically, the pseudo augmentation will be applied with the probability $p$ or be skipped with the probability $1-p$.

We note that the overfitting state of $D$ is dynamic throughout the training~(see Figure~\ref{fig:doverfit}).
It is intuitive to let the deception probability $p$ be adjusted adaptively based on the degree of overfitting. Ideally, $p$ should be adjusted without manual tuning regardless of data scales and properties.
To achieve this goal, inspired by ADA~\cite{ada}, we apply an overfitting heuristic $\lambda$ that quantifies the degree of $D$'s overfitting. We extend the control scheme of ADA and provide three plausible variants:
\begin{equation}
\small
\label{eq:3}
  \lambda_{r}=\mathbb{E}\left(\mathrm{sign}\left(D_{\mathrm{real}}\right)\right), \quad
  \lambda_{f}=-\mathbb{E}\left(\mathrm{sign}\left(D_{\mathrm{fake}}\right)\right), \quad
  \lambda_{rf}=\frac{\mathbb{E}\left(\mathrm{sign}\left(D_{\mathrm{real}}\right)\right)-\mathbb{E}\left(\mathrm{sign}\left(D_{\mathrm{fake}}\right)\right)}{2},
\end{equation}
where $D_{\mathrm{real}}$ and $D_{\mathrm{fake}}$ are defined as
\begin{equation}
\label{eq:2}
  D_{\mathrm{real}}=\mathrm{logit}\left(D\left(x\right)\right), \quad D_{\mathrm{fake}}=\mathrm{logit}\left(D\left(G\left(z\right)\right)\right),
\end{equation}
where $\mathrm{logit}$ denotes the logit function.
As shown in Figure~\ref{fig:doverfit}, the $\lambda_{r}$ in Eq.~\eqref{eq:3} estimates the portion of real images that obtain positive logit predictions by $D$, and that of generated images is captured by $\lambda_{f}$. Besides, $\lambda_{rf}$ indicates half of the distance between the signs of the real and fake logits.
For all these heuristics, $\lambda=0$ represents no overfitting, and $\lambda=1$ means complete overfitting. We use $\lambda_{r}$ in our main experiments and study other variants in the ablation study.

The strategy of using $\lambda$ to adjust $p$ is as follows.
We set a threshold $t$ (in most cases of our experiments, $t=0.6$) and initialize $p$ to be zero. If $\lambda$ signifies too much/little overfitting regarding $t$ (\ie, larger/smaller than $t$), the probability $p$ will be increased/decreased by one fixed step. Using this step size, $p$ can increase from zero to one in $500$k images shown to $D$. We adjust $p$ once every four iterations and clamp $p$ from below to zero after each adjustment.
In this way, the strength of pseudo augmentation can be adaptively controlled based on the degree of overfitting (see Figure~\ref{fig:mainidea}).

\subsection{Theoretical Analysis}
\label{sec:theory}

Let $p_{z}\left(z\right)$ be the prior on the input noise variable. The mapping from the latent space to the image space is denoted as $G\left(z\right)$. For sample $x$, $D\left(x\right)$ represents the estimated probability of $x$ coming from the real data.
To examine the rationality of APA, we analyze it in a non-parametric setting~\cite{GAN}, where a model is represented with infinite capacity by exploring its convergence in the space of probability density functions. Ideally, the estimated probability distribution $p_g$ defined by $G$ should  perfectly model the real data distribution $p_\mathrm{data}$ without bias if given enough capability and training time.

Since the deception strength $p$ is adaptively adjusted, to facilitate this theoretical analysis, we make a mild assumption that $\alpha$ is the expected strength that approximates the effect of dynamic adjustment during the entire training procedure.
Since $p\in\left[0,1\right)$, we have $0 \leq \alpha \textless p_{\mathrm{max}} \textless 1$, where $p_{\mathrm{max}}$ is the maximum of deception strength throughout training.
Hence, the value function $V\left(G, D\right)$ for the minimax two-player game of APA can be reformulated as:
\begin{equation}
\label{eq:4}
\begin{aligned}
  \min_{G}{\max_{D}{V\left(G, D\right)}}=
  \left(1-\alpha\right) \mathbb{E}_{x \sim p_{\mathrm{data}}\left(x\right)}\left[\log{D\left(x\right)}\right]
  &+\alpha\ \mathbb{E}_{z \sim p_{z}\left(z\right)}\left[\log{D\left(G\left(z\right)\right)}\right]\\
  &+\mathbb{E}_{z \sim p_{z}\left(z\right)}\left[\log{\left(1-D\left(G\left(z\right)\right)\right)}\right],
\end{aligned}
\end{equation}

First, let us consider the optimal discriminator~\cite{GAN} for any given generator.

\textbf{Proposition 1.} \textit{If the generator $G$ is fixed, the optimal discriminator $D$ for APA is:}
\begin{equation}
\label{eq:5}
  D_{G}^{*}\left(x\right)=
  \frac
  {\left(1-\alpha\right)p_\mathrm{data}\left(x\right)+\alpha\ p_g\left(x\right)}
  {\left(1-\alpha\right)p_\mathrm{data}\left(x\right)+\left(1+\alpha\right)p_g\left(x\right)}
\end{equation}

\textit{Proof.} Applying APA, given any generator $G$, the training objective of the discriminator $D$ is to maximize the value function $V\left(G, D\right)$ in Eq.~\eqref{eq:4}:
\begin{equation}
\small
\label{eq:6}
\begin{aligned}
  V\left(G, D\right)
  &=\left(1-\alpha\right)\int_{x}{p_{\mathrm{data}}\left(x\right)\log{D\left(x\right)}dx}
  +\alpha\int_{z}{p_{z}\left(z\right)\log{D\left(g\left(z\right)\right)}dz}
  +\int_{z}{p_{z}\left(z\right)\log{\left(1-D\left(g\left(z\right)\right)\right)}dz}\\
  &=\int_{x}{\left[
  \left(1-\alpha\right)p_{\mathrm{data}}\left(x\right)\log{D\left(x\right)}
  +\alpha\ p_{g}\left(x\right)\log{D\left(x\right)}
  +p_{g}\left(x\right)\log{\left(1-D\left(x\right)\right)}
  \right]dx}\\
  &=\int_{x}{\left[
  \left(\left(1-\alpha\right)p_{\mathrm{data}}\left(x\right)+\alpha\ p_{g}\left(x\right)\right)\log{D\left(x\right)}
  +p_{g}\left(x\right)\log{\left(1-D\left(x\right)\right)}
  \right]dx}
\end{aligned}
\end{equation}
For any $\left(m, n\right)\in\mathbb{R}^2\ \backslash\left\{0, 0\right\}$, the function $f\left(y\right)=m\log{\left(y\right)}+n\log{\left(1-y\right)}$ achieves its maximum in the range $\left[0, 1\right]$ at $\frac{m}{m+n}$.
Besides, the discriminator $D$ is defined only inside of $\mathrm{supp}\left(p_{\mathrm{data}}\right)\cup\mathrm{supp}\left(p_{g}\right)$, where $\mathrm{supp}$ is the set-theoretic support.
Therefore, we conclude the proof for Proposition~1.

We have got the optimal discriminator $D_{G}^{*}\left(x\right)$ in Eq.~\eqref{eq:5} that maximizes the value function $V\left(G, D\right)$ given any fixed generator $G$. The goal of generator $G$ in the adversarial training is to minimize the value function $V\left(G, D\right)$ in Eq.~\eqref{eq:4} when $D$ achieves the optimum.
Since the training objective of $D$ can be interpreted as maximizing the log-likelihood for the conditional probability $P\left(Y=y|x\right)$, where $Y$ estimates that $x$ comes from $p_{\mathrm{data}}$ (\ie, $y=1$) or from $p_{g}$ (\ie, $y=0$), we reformulate virtual training criterion~\cite{GAN} as:
\begin{equation}
\small
\label{eq:7}
\begin{aligned}
  C\left(G\right)
  &=\left(1-\alpha\right)\mathbb{E}_{x \sim p_{\mathrm{data}}}\left[\log{D_{G}^{*}\left(x\right)}\right]
  +\alpha\ \mathbb{E}_{z \sim p_{z}}\left[\log{D_{G}^{*}\left(G\left(z\right)\right)}\right]
  +\mathbb{E}_{z \sim p_{z}}\left[\log{\left(1-D_{G}^{*}\left(G\left(z\right)\right)\right)}\right]\\
  &=\left(1-\alpha\right)\mathbb{E}_{x \sim p_{\mathrm{data}}}\left[\log{D_{G}^{*}\left(x\right)}\right]
  +\alpha\ \mathbb{E}_{x \sim p_{g}}\left[\log{D_{G}^{*}\left(x\right)}\right]
  +\mathbb{E}_{x \sim p_{g}}\left[\log{\left(1-D_{G}^{*}\left(x\right)\right)}\right]
\end{aligned}
\end{equation}

Then, let us consider the global minimum of $C\left(G\right)$ trained with the proposed APA.

\textbf{Proposition 2.} \textit{Applying APA, the global minimum of the virtual training criterion $C\left(G\right)$ is still achieved if and only if $p_{g}=p_{\mathrm{data}}$, where $C\left(G\right)=-\log{4}$.}

\textit{Proof.}
1) If $p_{g}=p_{\mathrm{data}}$, we have $D_{G}^{*}\left(x\right)=\frac{1}{2}$ according to Eq.~\eqref{eq:5}. By inspecting Eq.~\eqref{eq:7} at $D_{G}^{*}\left(x\right)=\frac{1}{2}$, we get $C^{*}\left(G\right)=\left(1-\alpha\right)\log{\frac{1}{2}}+\alpha\log{\frac{1}{2}}+\log{\frac{1}{2}}=\log{\frac{1}{2}}+\log{\frac{1}{2}}=-\log{4}$.

2) To verify $C^{*}\left(G\right)$ is the global minimum of $C\left(G\right)$, and it can only be achieved when $p_{g}=p_{\mathrm{data}}$, as in the derivation of Eq.~\eqref{eq:6}, we obtain:
\begin{equation}
\label{eq:8}
\begin{aligned}
  C\left(G\right)
  =\int_{x}{\left(\left(1-\alpha\right)p_{\mathrm{data}}\left(x\right)+\alpha\ p_{g}\left(x\right)\right)\log{D_{G}^{*}\left(x\right)}dx}
  +\int_{x}{p_{g}\left(x\right)\log{\left(1-D_{G}^{*}\left(x\right)\right)}dx}
\end{aligned}
\end{equation}
Observe that
\begin{equation}
\label{eq:9}
\begin{aligned}
  -\log{4}
  &=\left(1-\alpha\right)\mathbb{E}_{x \sim p_{\mathrm{data}}}\left[-\log{2}\right]
  +\alpha\ \mathbb{E}_{x \sim p_{g}}\left[-\log{2}\right]
  +\mathbb{E}_{x \sim p_{g}}\left[-\log{2}\right]\\
  &=-\int_{x}{\left(\left(1-\alpha\right)p_{\mathrm{data}}\left(x\right)+\alpha\ p_{g}\left(x\right)\right)\log{2}\ dx}
  -\int_{x}{p_{g}\left(x\right)\log{2}\ dx}
\end{aligned}
\end{equation}
Subtracting Eq.~\eqref{eq:9} from Eq.~\eqref{eq:8},
\begin{equation}
\small
\label{eq:10}
\begin{aligned}
  C\left(G\right)
  =-\log{4}+\int_{x}{\left(\left(1-\alpha\right)p_{\mathrm{data}}\left(x\right)+\alpha\ p_{g}\left(x\right)\right)\log{2\cdot D_{G}^{*}\left(x\right)}dx}
  +\int_{x}{p_{g}\left(x\right)\log{2\cdot \left(1-D_{G}^{*}\left(x\right)\right)}dx}
\end{aligned}
\end{equation}
By substituting Eq.~\eqref{eq:5} into Eq.~\eqref{eq:10}, we achieve:
\begin{equation}
\label{eq:11}
\begin{aligned}
  C\left(G\right)
  =-\log{4}
  &+\mathrm{KLD}\left(\left(\left(1-\alpha\right)p_{\mathrm{data}}+\alpha\ p_{g}\right)
  \bigg\|\frac{\left(1-\alpha\right)p_\mathrm{data}+\left(1+\alpha\right)p_g}{2}\right)\\
  &+\mathrm{KLD}\left(p_{g}
  \bigg\|\frac{\left(1-\alpha\right)p_\mathrm{data}+\left(1+\alpha\right)p_g}{2}\right),
\end{aligned}
\end{equation}
where $\mathrm{KLD}$ is the Kullback-Leibler (KL) divergence. Moreover, Eq.~\eqref{eq:11} further implies that the generation process of $G$ by APA can be regarded as minimizing the Jensen-Shannon (JS) divergence between the smoothed data distribution and the generated distribution:
\begin{equation}
\label{eq:12}
\begin{aligned}
  C\left(G\right)
  =-\log{4}
  &+2\cdot \mathrm{JSD}\left(\left(\left(1-\alpha\right)p_{\mathrm{data}}+\alpha\ p_{g}\right)
  \|p_{g}\right).
\end{aligned}
\end{equation}
For the two distributions $P$ and $Q$, their JS divergence $\mathrm{JSD}\left(P\|Q\right)\geq0$ and $\mathrm{JSD}\left(P\|Q\right)=0$ if and only if $P=Q$. Therefore, for $0 \leq \alpha \textless p_{\mathrm{max}} \textless 1$, we obtain that $C^{*}\left(G\right)=-\log{4}$ is the global minimum of $C\left(G\right)$, and the only solution is $\left(1-\alpha\right)p_{\mathrm{data}}+\alpha\ p_{g}=p_{g}$, \ie, $p_{g}=p_{\mathrm{data}}$. Q.E.D.

Given the proof in~\cite{GAN}, if $G$ and $D$ have enough capacity to reach their optimum, Proposition~2 indicates that the generated distribution $p_g$ can ideally converge to the real data distribution $p_{\mathrm{data}}$.
So far, we have proved the convergence of $G$ trained with our proposed APA, which can perfectly model the real data distribution given sufficient capability and training time.
Besides, the JS divergence term between the smoothed data distribution and the generated distribution in Eq.~\eqref{eq:12} implies that the judgment of $D$ may be moderated to alleviate overfitting.
These conclusions explain the rationality of the proposed APA for training GANs with limited data.


\section{Experiments}
\label{sec:experiments}

\textbf{Datasets.}
We use four datasets in our main experiments: Flickr-Faces-HQ (FFHQ)~\cite{stylegan} with $70,000$ human face images, AFHQ-Cat~\cite{starganv2} with $5,153$ cat faces, Caltech-UCSD Birds-200-2011 (CUB)~\cite{cub2002011} with $11,788$ images of birds, and Danbooru2019 Portraits (Anime)~\cite{danbooru2019portraits} with $302,652$ anime portraits.
We exploit some of their artificially limited subsets under different settings.
All the images are resized to a moderate resolution of $256 \times 256$ using a high-quality Lanczos filter~\cite{Lanczos} to reduce the energy consumption for large-scale GAN training while preserving image quality.
Additional dataset details, including the data source and license information, can be found in the \textit{Appendix}.

\textbf{Evaluation metrics.}
We follow the standard evaluation protocol~\cite{BigGAN,reglimited} for the quantitative evaluation.
Specifically,
we use the Fr$\mathrm{\acute{e}}$chet Inception Distance (FID, lower is better)~\cite{TTUR}, which quantifies the distance between distributions for the real and generated images. FID evaluates the realness of synthesized images. Following~\cite{TTUR,ada}, we calculate FID for the models trained with limited datasets using $50$k generated images and all the real images in the original datasets.
We also apply the Inception Score (IS, higher is better)~\cite{improvedtechgans}. IS considers the clarity and diversity of generated images.

\textbf{Implementation details.}
We choose the state-of-the-art StyleGAN2~\cite{stylegan2} as the backbone to verify the effectiveness of APA on limited data.
We use the default setups of APA provided in Section~\ref{sec:apa} unless specified otherwise.
For a fair and controllable comparison, we reimplement all baselines and run the experiments from scratch using official code.
All the models are trained on 8 NVIDIA Tesla V100 GPUs.
Please refer to the \textit{Appendix} for more implementation details and additional benchmark results (\eg, performance on BigGAN~\cite{BigGAN}).

\subsection{The Effectiveness of APA}
\label{sec:expeffectiveness}

\begin{figure}
	\centering
	\includegraphics[width=\linewidth]{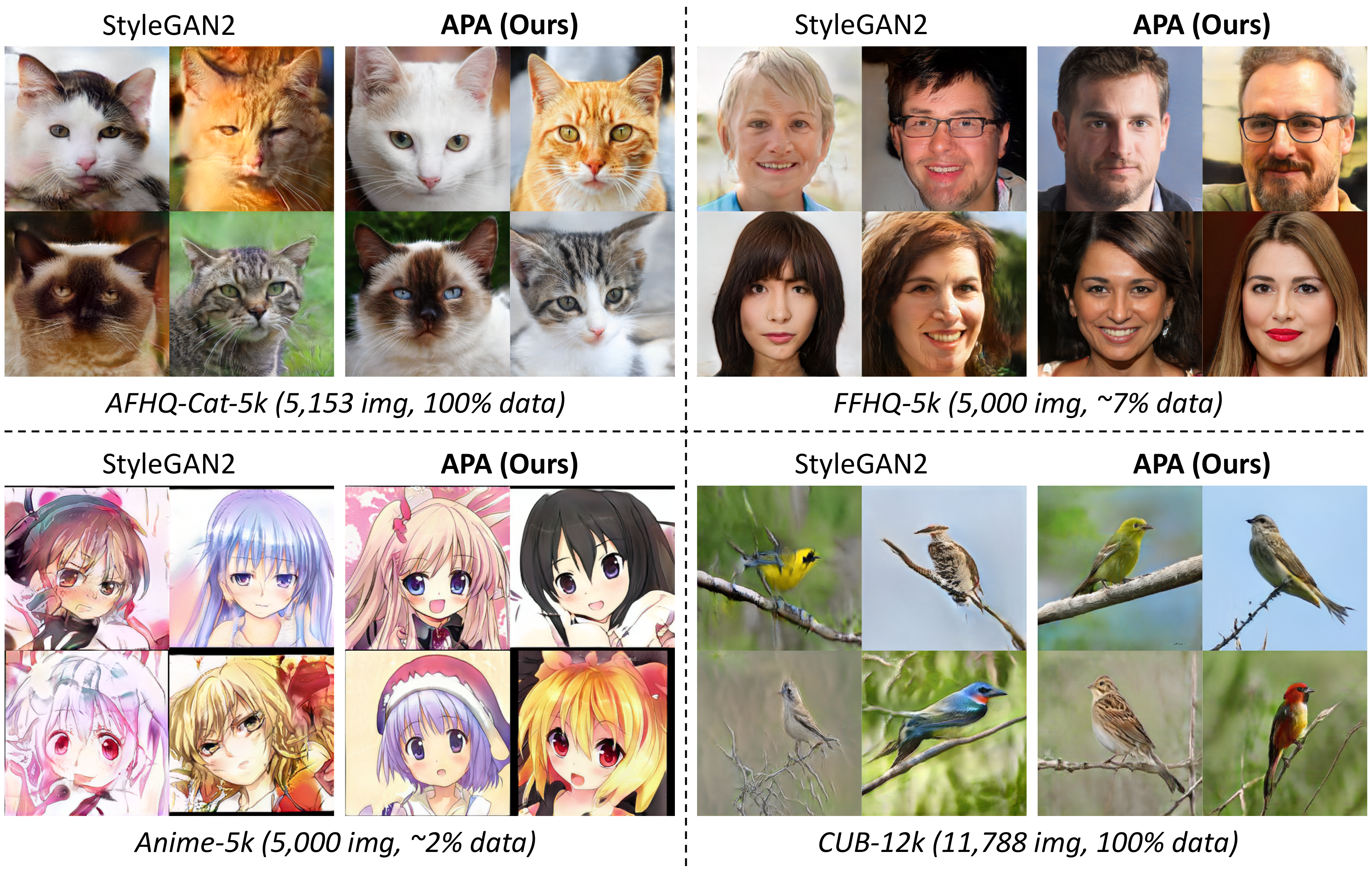}
	\caption{The proposed APA improves StyleGAN2~\cite{stylegan2} synthesized results ($256 \times 256$, no truncation) on \textbf{various datasets} with limited data amounts. We randomly select subsets to confine the size of large datasets (\ie, FFHQ-5k~\cite{stylegan} and Anime-5k~\cite{danbooru2019portraits}) and directly use small datasets (\ie, AFHQ-Cat-5k~\cite{starganv2} and CUB-12k~\cite{cub2002011}) whose data amount is already insufficient for StyleGAN2.}
	\label{fig:effectonsg2}
\end{figure}

\begin{table}[tb!]
\centering
\small
\caption{The FID (lower is better) and IS (higher is better) scores ($256 \times 256$) of our method compared to state-of-the-art StyleGAN2 on \textbf{various datasets} with limited data amounts.}
\begin{tabularx}{\textwidth}{l|*{8}{|Y}}
\Xhline{1pt}
& \multicolumn{2}{c}{AFHQ-Cat-5k} & \multicolumn{2}{|c}{FFHQ-5k} & \multicolumn{2}{|c}{Anime-5k}& \multicolumn{2}{|c}{CUB-12k} \\
\cline{2-9}
Method& FID $\downarrow$& IS $\uparrow$& FID $\downarrow$& IS $\uparrow$& FID $\downarrow$& IS $\uparrow$& FID $\downarrow$& IS $\uparrow$ \\
\Xhline{0.6pt}
StyleGAN2~\cite{stylegan2} & 7.737& 1.825& 37.830& 4.018& 23.778& 2.289& 23.437& 5.812 \\
APA (Ours)& {\bf4.876}& {\bf2.156}& {\bf13.249}& {\bf4.487}& {\bf13.089}& {\bf2.330}& {\bf12.889}& {\bf5.869} \\
\Xhline{1pt}
\end{tabularx}
\label{tbl:effectonsg2}
\vspace{-0.31cm}
\end{table}

\textbf{Effectiveness on various datasets.}
The comparative results of StyleGAN2 on various datasets with limited data amounts are shown in Figure~\ref{fig:effectonsg2}.
The quality of images synthesized by StyleGAN2 deteriorates under limited data. Ripple artifacts appear on the cat faces and human faces, and the facial features of the anime faces are misplaced. On the bird dataset with heavy background clutter, the generated images are completely distorted albeit trained with more data.
The proposed APA significantly ameliorates image quality on all these datasets, producing much more photorealistic results.
The quantitative evaluation results are reported in Table~\ref{tbl:effectonsg2}. Applying APA contributes to a performance boost of FID and IS in all cases, suggesting that the synthesized images are with higher quality and diversity on different datasets.

\begin{figure}
	\centering
	\includegraphics[width=\linewidth]{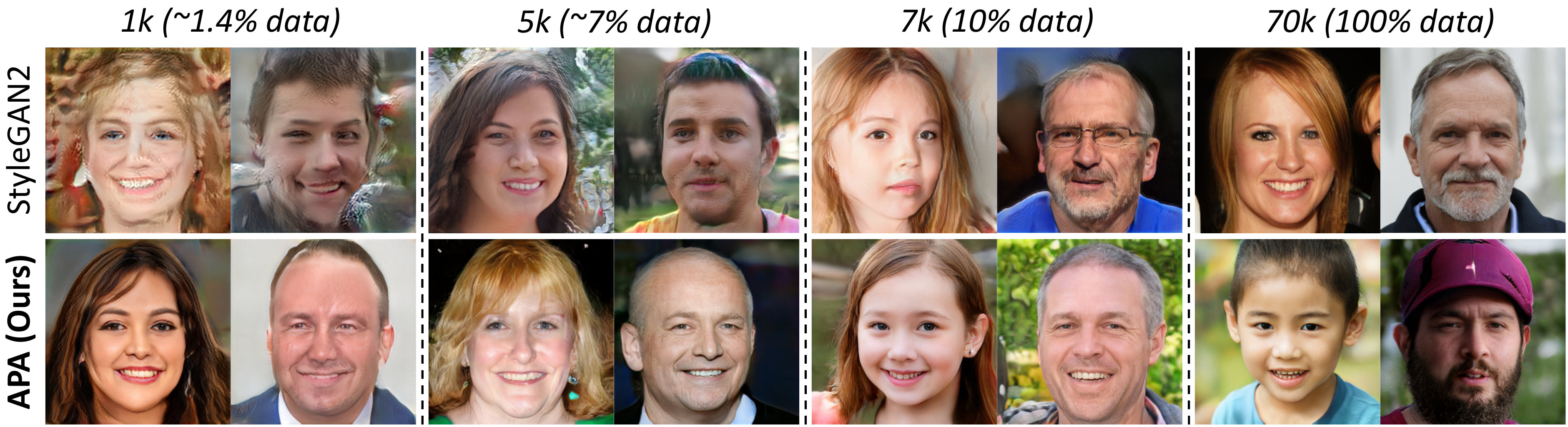}
	\caption{The effectiveness of APA to improve StyleGAN2~\cite{stylegan2} synthesized results ($256 \times 256$, no truncation) on the subsets of FFHQ~\cite{stylegan} with \textbf{different data amounts}.}
	\label{fig:ffhqdiffamount}
\end{figure}

\textbf{Effectiveness given different data amounts.}
The comparative results on subsets of FFHQ~\cite{stylegan}, with varying amounts of data, are shown in Figure~\ref{fig:ffhqdiffamount}. The corresponding quantitative test results are presented in Table~\ref{tbl:ffhqdiffamount}.
APA improves the image quality and metric performance in all cases. Notably, the quality of synthesized images by APA on 5k/7k data is visually close to StyleGAN2 results on the full dataset while with an order of magnitude fewer training samples.
As for the quantitative results, the IS score of APA on 1k data is better than that of StyleGAN2 on 5k data, and both metrics of APA on 5k data outperform StyleGAN2 results on 7k data. APA can even improve StyleGAN2 performance on the full dataset, further indicating its potential.

\begin{table}[tb!]
\centering
\small
\caption{The FID (lower is better) and IS (higher is better) scores ($256 \times 256$) of our method on StyleGAN2 trained using the subsets of FFHQ~\cite{stylegan} with \textbf{different data amounts}.}
\begin{tabularx}{\textwidth}{l|*{8}{|Y}}
\Xhline{1pt}
& \multicolumn{2}{c}{1k ($\sim1.4\%$)} & \multicolumn{2}{|c}{5k ($\sim7\%$)} & \multicolumn{2}{|c}{7k ($10\%$)}& \multicolumn{2}{|c}{70k ($100\%$)} \\
\cline{2-9}
Method& FID $\downarrow$& IS $\uparrow$& FID $\downarrow$& IS $\uparrow$& FID $\downarrow$& IS $\uparrow$& FID $\downarrow$& IS $\uparrow$ \\
\Xhline{0.6pt}
StyleGAN2~\cite{stylegan2} & 86.407& 2.806& 37.830& 4.018& 27.738& 4.264& 3.862& 5.243 \\
APA (Ours)& {\bf45.192}& {\bf4.130}& {\bf13.249}& {\bf4.487}& {\bf10.800} & {\bf4.860} & {\bf3.678}& {\bf5.336} \\
\Xhline{1pt}
\end{tabularx}
\label{tbl:ffhqdiffamount}
\vspace{-0.35cm}
\end{table}

\begin{figure}
	\centering
	\includegraphics[width=\linewidth]{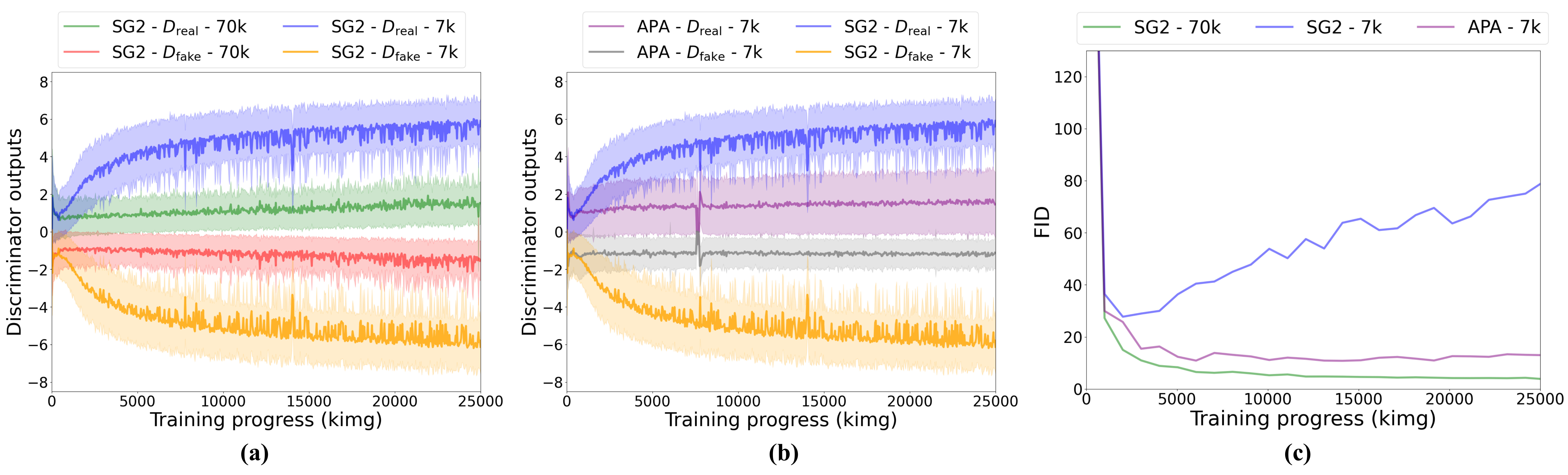}
	\caption{The \textbf{overfitting and convergence status} of APA compared to StyleGAN2 (SG2) on FFHQ~\cite{stylegan} ($256 \times 256$). (a) The discriminator raw output logits of StyleGAN2 on the full (70k) or limited (7k) datasets. (b) The discriminator raw output logits of StyleGAN2 and APA on the limited (7k) dataset. (c) The training convergence shown by FID.}
	\label{fig:overfitsg2apa}
\end{figure}

\textbf{Overfitting and convergence analysis.}
As shown in Figure~\ref{fig:overfitsg2apa}, the divergence of StyleGAN2 discriminator predictions can be effectively restricted on FFHQ-7k ($10\%$ of full data) by applying the proposed APA. The curves of APA on FFHQ-7k become closer to that of StyleGAN2 on FFHQ-70k, suggesting the effectiveness of APA in curbing the overfitting of the discriminator.
Besides, APA improves the training convergence of StyleGAN2 on limited data, shown by the FID curves.
More overfitting and convergence analysis can be found in the \textit{Appendix}.

\subsection{Comparison with Other Solutions for GAN Training with Limited Data}
\label{sec:expcompare}

\textbf{Performance and compatibility.}
We compare the proposed APA with representative approaches designed for the low-data regime, including ADA~\cite{ada} and LC-regularization (LC-Reg)~\cite{reglimited}, which perform standard data augmentations and model regularization, respectively.
The results are reported in Table~\ref{tbl:compare}. As a single method, APA outperforms previous solutions in most cases, effectively improving the StyleGAN2 baseline on both the limited and full datasets. Although ADA~\cite{ada} achieves slightly better results than our method on FFHQ-5k, it yields a worse FID score with StyleGAN2 on the full dataset. Applying LC-Reg~\cite{reglimited} needs careful manual tuning, and its own effect on limited data is not apparent compared to other methods.

It is noteworthy that APA is also complementary to existing methods based on standard data augmentations, \eg, ADA~\cite{ada}. As can be observed in Table~\ref{tbl:compare}, APA can further boost the performance of StyleGAN2 with ADA~\cite{ada} given limited training data, suggesting the compatibility of our approach with standard data augmentations.
Combining ADA~\cite{ada} and APA on FFHQ full data outperforms StyleGAN2 but is slightly inferior to applying APA solely. The degraded performance is mainly affected by ADA~\cite{ada}, which we empirically found might slightly harm the performance when the training data is sufficient. Overall, these methods are not dedicated to improving performance under sufficient data. Nevertheless, as a beneficial side effect, the proposed APA may have this potential.
More comparative results are included in the \textit{Appendix}.

\begin{table}[tb!]
\centering
\small
\caption{The FID (lower is better) and IS (higher is better) scores ($256 \times 256$) of our method \textbf{compared to other state-of-the-art solutions designed for GAN training with limited data} on StyleGAN2. The bold number indicates the best value, and the underline marks the second best.}
\begin{tabularx}{\textwidth}{l|*{6}{|Y}}
\Xhline{1pt}
& \multicolumn{2}{c}{AFHQ-Cat-5k} & \multicolumn{2}{|c}{FFHQ-5k} & \multicolumn{2}{|c}{FFHQ-70k (full)} \\
\cline{2-7}
Method& FID $\downarrow$& IS $\uparrow$& FID $\downarrow$& IS $\uparrow$& FID $\downarrow$& IS $\uparrow$ \\
\Xhline{0.6pt}
StyleGAN2~\cite{stylegan2} & 7.737& 1.825& 37.830& 4.018& \underline{3.862}& 5.243 \\
ADA~\cite{ada}& 6.053& 2.119& \underline{11.409}& \underline{4.721}& 4.018& \underline{5.329} \\
LC-Reg~\cite{reglimited}& 6.699& 1.943& 35.148& 3.926& 3.933 & 5.312 \\
\cline{1-7}
APA (Ours)& \underline{4.876}& \underline{2.156}& 13.249& 4.487& \bf{3.678}& \bf{5.336} \\
ADA + APA (Ours)& {\bf4.377}& {\bf2.169}& {\bf8.379}& {\bf4.849}& 3.811& 5.321 \\
\Xhline{1pt}
\end{tabularx}
\label{tbl:compare}
\end{table}

\textbf{Training cost.}
We compare the computational cost of APA against ADA~\cite{ada}, using the same basic official codebase.
There is no parameter or memory increment for both methods.
As for the time consumption, we test the training cost on 8 NVIDIA Tesla V100 GPUs. On FFHQ-5k~($256 \times 256$), the average training time of the StyleGAN2~\cite{stylegan2} baseline is ($4.740\pm0.100$) sec/kimg (\ie, seconds per thousand of images shown to the discriminator). The cost of our method is negligible, slightly increasing this value to ($4.789\pm0.078$) sec/kimg. As a reference, the value for ADA~\cite{ada} is ($5.327\pm0.116$) sec/kimg, spending additional time for applying external augmentations.

\subsection{Comparison with Previous Techniques for Regularizing GANs}
\label{sec:expcompareganreg}

\begin{table}[tb!]
\centering
\small
\caption{The FID (lower is better) and IS (higher is better) scores ($256 \times 256$) of our method \textbf{compared to previous techniques for regularizing GANs} on StyleGAN2 trained with FFHQ-5k~\cite{stylegan}.}
\begin{tabularx}{\textwidth}{l|*{4}{|Y}}
\Xhline{1pt}
\cline{2-5}
Metric& StyleGAN2~\cite{stylegan2}& Instance noise~\cite{noisestable2}& One-sided LS~\cite{improvedtechgans}& APA (Ours) \\
\Xhline{0.6pt}
FID $\downarrow$ & 37.830& 40.981& 33.978& {\bf13.249} \\
IS $\uparrow$ & 4.018& 4.231& 4.029& {\bf4.487} \\
\Xhline{1pt}
\end{tabularx}
\label{tbl:compareganreg}
\vspace{-0.3cm}
\end{table}

As mentioned in Section~\ref{sec:relatedwork}, APA is closely related to previous techniques for regularizing GANs. The comparative results on APA and some representative conventional techniques, \ie, instance noise~\cite{noisestable2} and one-sided label smoothing (LS)~\cite{improvedtechgans}, are shown in Table~\ref{tbl:compareganreg}.
Applying instance noise~\cite{noisestable2} may not boost the performance of StyleGAN2 much under limited data. One-sided label smoothing~(LS)~\cite{improvedtechgans} (with the real label of $0.9$) outperforms StyleGAN2 but still has a huge performance gap with our method. This further suggests the effectiveness and usefulness of APA.

\subsection{Ablation Studies}
\label{sec:expablation}

\textbf{Ablation studies on variants of APA.}
We study three key elements of APA, \ie, the overfitting heuristic $\lambda$, the deception strength $p$, and the deception strategy.
The version used in our main experiments is denoted as the ``main'' version, where $\lambda=\lambda_{r}$, and $p$ is adjusted adaptively. Besides, the ``main'' version applies the deception strategy that is analogous to one-sided label flipping.
As reported in Table~\ref{tbl:ablation}, when using other variants of $\lambda$ we suggested in Eq.~\eqref{eq:3} (\ie, $\lambda=\lambda_{f}$ and $\lambda=\lambda_{rf}$), the models achieve comparable performance as the main version. The FID score becomes even better for $\lambda=\lambda_{rf}$, indicating the flexibility of our provided heuristics to extend and modify APA.
More interestingly, even a fixed moderate deception probability (\eg, $p=0.5$) can still work much better than original StyleGAN2 on limited data, albeit slightly inferior to the adaptively adjusted $p$. This implies the importance of the pseudo augmentation, and the adaptive control scheme can further boost performance without manual tuning.
As for the deception strategy, we empirically observe that two-sided label flipping can still outperform StyleGAN2 but is inferior to the main version.

\begin{table}[tb!]
\centering
\small
\caption{\textbf{Ablation studies on variants of APA} on FFHQ-5k~\cite{stylegan} ($256 \times 256$). We study three key elements of APA, \ie, the overfitting heuristic $\lambda$, the deception strength $p$, and the deception strategy. The ``main'' denotes the main version used in our previous experiments (\ie, $\lambda=\lambda_{r}$, $p$ is adaptively adjusted, and the deception strategy is analogous to one-sided label flipping).}
\begin{tabularx}{\textwidth}{l|*{6}{|Y}}
\Xhline{1pt}
\cline{2-7}
Metric& StyleGAN2& main& $\lambda=\lambda_f$& $\lambda=\lambda_{rf}$& $p=0.5$ (fix)& two-sided  \\
\Xhline{0.6pt}
FID $\downarrow$ & 37.830& \underline{13.249}& 13.470& {\bf12.679}& 14.632& 15.440 \\
IS $\uparrow$ & 4.018& {\bf4.487}& \underline{4.420}& 4.412& 4.403& 4.167 \\
\Xhline{1pt}
\end{tabularx}
\label{tbl:ablation}
\vspace{-0.2cm}
\end{table}

\textbf{Ablation studies on the threshold $\bm t$.}
We further provide the ablation studies on the threshold value~$t$ in Table~\ref{tbl:ablationt}. The version used in our main experiments is denoted as the ``main'' version, where $t=0.6$. It can be seen that the models with different values of $t$ achieve comparable results, outperforming the StyleGAN2 baseline. On FFHQ-5k ($256 \times 256$), $t=0.6$ could be a more plausible choice.
For a further explanation, we use $t=0.6$ as the default value since it works well in most cases. In practice, the value of $t$ could be further adjusted to achieve even better results. Empirically, a smaller $t$ can be chosen when one has fewer data. This means the deception strength $p$ can be adjusted to increase more rapidly since the discriminator is more prone to overfitting when the data amount is fewer.

\begin{table}[tb!]
\centering
\small
\caption{\textbf{Ablation studies on the threshold $\bm t$} on FFHQ-5k~\cite{stylegan} ($256 \times 256$). The ``main'' denotes the main version used in our previous experiments (\ie, $t=0.6$).}
\begin{tabularx}{\textwidth}{l|*{4}{|Y}}
\Xhline{1pt}
\cline{2-5}
Metric& StyleGAN2& $t=0.4$& $t=0.6$ (main)& $t=0.8$ \\
\Xhline{0.6pt}
FID $\downarrow$ & 37.830& 13.687& {\bf13.249}& 13.984 \\
IS $\uparrow$ & 4.018& 4.418& {\bf4.487}& 4.395 \\
\Xhline{1pt}
\end{tabularx}
\label{tbl:ablationt}
\vspace{-0.55cm}
\end{table}


\section{Discussion}
\label{sec:discussion}

We have shown the effectiveness of the proposed adaptive pseudo augmentation (APA) for state-of-the-art GAN training with limited data empirically.
With negligible computational cost, APA achieves comparable or even better performance than other types of solutions on various datasets. APA is also complementary to existing methods based on standard data augmentations.

\textbf{Limitations.}
Despite promising results, the quality of synthesized images by APA on the datasets with extremely limited data amount (\eg, hundreds of images) can still be improved.
Besides, on certain datasets such as FFHQ-5k, applying APA solely may be slightly inferior to approaches based on standard data augmentations.
Since we do not apply any external augmentations, these two limitations are both due to the insufficiency of the dataset's intrinsic diversity.
These limitations may be approached in the future in two ways:
1) Incorporating better standard data augmentations to APA.
2) Exploring the issue of data insufficiency from the generator aspect, \eg, using a multi-modal generator~\cite{multimodalg} to enhance diversity.
In addition, we only theoretically verified the convergence and rationality of APA. In future work, the theoretical analysis on the effectiveness of APA could be further explored.

\textbf{Broader impact.}
On the one hand, the effectiveness of APA with negligible computational cost will benefit the practical deployment of GANs, especially in the low-data regime. APA may also extend the breadth and potential of solutions to training GANs with limited data and benefit downstream tasks, such as conditional synthesis.
On the other hand, APA may also bring potential concerns on its capability to ease the higher-quality fake media synthesis using only limited data, as technology is usually a double-edged sword. However, we believe that these concerns can be resolved by developing better media forensics methods and datasets as countermeasures.

\textbf{Acknowledgments and funding disclosure.}
This study is supported under the RIE2020 Industry Alignment Fund – Industry Collaboration Projects (IAF-ICP) Funding Initiative, as well as cash and in-kind contribution from the industry partner(s).

{\small
\bibliographystyle{plain}
\bibliography{sections_arxiv/egbib}
}


\section*{Appendix}
\label{sec:appendix}
This appendix provides supplementary information that is not elaborated in our main paper:
Section~\ref{sec:preliminary} describes some preliminary concepts and definitions of our methodology.
Section~\ref{sec:discussganreg} provides additional discussions on previous techniques for regularizing GANs.
Section~\ref{sec:dataset} details our used datasets.
Section~\ref{sec:implementation} presents the implementation details of our experiments.
Section~\ref{sec:examples} shows more results and analysis.

\appendix
\section{Preliminaries of Methodology}
\label{sec:preliminary}

To further facilitate readers' understanding of the theoretical analysis and the proposed Adaptive Pseudo Augmentation (APA), this section will provide some preliminary concepts and definitions for the methodology section in our main paper.

Generative adversarial networks (GANs)~\cite{GAN} aim at capturing the real data distribution to synthesize new data.
Two networks are trained alternately via an adversarial process: a generator $G$ learns to produce new samples, and a discriminator $D$ (\ie, a binary classifier) predicts the probability that a sample comes from the real data rather than from $G$.
Following~\cite{GAN,stylegan2}, our main paper focuses on the fundamental problem of GANs, \ie, unconditional image synthesis, which is generating random samples from a noise input in the latent space. The noise is sampled from a Gaussian distribution.

The goal of GANs is to learn an ideal generated distribution $p_{g}$ from the real data distribution $p_{\mathrm{data}}$. Let $p_{z}\left(z\right)$ be the prior on the input noise variable. The mapping from the latent space to the image space can be denoted as $G\left(z\right)$. For sample $x$, $D\left(x\right)$ represents the estimated probability of $x$ coming from the real data. Here, both $G$ and $D$ should be differentiable functions that are defined by the network parameters.
To quantify the adversarial process, $G$ and $D$ play a minimax two-player game with the value function $V\left(G, D\right)$:
\begin{equation}
\label{eq:1}
  \min_{G}{\max_{D}{V\left(G, D\right)}}=\mathbb{E}_{x \sim p_{\mathrm{data}}\left(x\right)}\left[\log{D\left(x\right)}\right]+\mathbb{E}_{z \sim p_{z}\left(z\right)}\left[\log{\left(1-D\left(G\left(z\right)\right)\right)}\right].
\end{equation}
Let the virtual training criterion~\cite{GAN} for the generator $G$ be $C\left(G\right)$. The global minimum of $C\left(G\right)$ is achieved if and only if $p_{g}=p_{\mathrm{data}}$, and the minimum value is $-\log{4}$, as proved by~\cite{GAN}. This indicates that GANs can perfectly model the real data distribution if given sufficient capacity and time.
In practice, we usually use a non-saturated form for $G$ and train it to maximize $\log{D\left(G\left(z\right)\right)}$ instead of minimizing $\log{\left(1-D\left(G\left(z\right)\right)\right)}$ to ensure a healthy gradient at the early training stage.

In all the figures of this work, we show raw output logits of $D$ before the last Sigmoid activation to better visualize its prediction confidence. Let $\mathrm{logit}$ denotes the logit function, we define:
\begin{equation}
\label{eq:2}
  D_{\mathrm{real}}=\mathrm{logit}\left(D\left(x\right)\right), \quad D_{\mathrm{fake}}=\mathrm{logit}\left(D\left(G\left(z\right)\right)\right).
\end{equation}

\section{Additional Discussions on Previous Techniques for Regularizing GANs}
\label{sec:discussganreg}

As mentioned in our main paper, previous techniques for regularizing GANs include adding noise to the inputs of the discriminator~\cite{noisestable1,noisestable2,noisestable3}, gradient penalty~\cite{r1reg,wgangp,stablegp}, one-sided label smoothing~\cite{improvedtechgans}, spectral normalization~\cite{spectralnorm}, label noise~\cite{nips16ganworkshop}, \etc.
These approaches are designed for stabilizing training or preventing mode collapse~\cite{improvedtechgans}. The essence of their objectives could be considered similar to our method since training GANs in the low-data regime exhibits similar behaviors as previously observed in early GANs with sufficient data.
However, several differences are worth highlighting.

First of all, the actual goals of previous strategies and the proposed APA may not be completely the same. Specifically, APA is specialized for training GANs in the low-data regime, which was not carefully considered by prior studies. This difference in the data setting is quite important since it is the core problem we wish to address.
Even the performance of state-of-the-art StyleGAN2~\cite{stylegan2} deteriorates when trained with a limited amount of data, although it has exploited many advanced techniques for stabilizing training or preventing mode collapse, such as R1 regularization~\cite{r1reg}.
The main challenge that lies within the low-data regime is the overfitting of the discriminator. Although this issue might also appear on early GANs, it becomes more severe when data is limited.

Besides, we have presented the comparative studies on APA and conventional techniques in the main paper. Empirically, we showed that previous methods could not boost the performance of StyleGAN2 much under limited data. Some of them showed a huge performance gap in comparison to the proposed APA.
The experiments indicate that previous relevant strategies cannot handle the low-data regime well, further suggesting the effectiveness and usefulness of APA.

Last but not least, APA and these techniques themselves are not exactly the same. The proposed APA is an effective practice and improvement of these ideas on modern GANs, whose implementations are very different from the early ones.
Compared to previous techniques, APA is more adaptive to fit different settings and the overfitting status in training. Although using adaptive heuristics was also explored in the past, it had been found unpractical at the time~\cite{nips16ganworkshop}. APA makes the adaptive control scheme possible in practice.

We believe that the proposed APA could contribute to the community for its effectiveness, simplicity, and adaptability for training state-of-the-art GANs in the low-data regime.
Hopefully, our approach could extend the breadth and potential of solutions to GAN training with limited data.

\section{Dataset Details}
\label{sec:dataset}

This section will detail our explored datasets in the main paper.
We randomly select subsets to confine the size of large datasets and directly use small datasets for GAN training with limited data under different settings.
Our main paper focuses on the fundamental unconditional image synthesis task with powerful contemporary GANs. Thus, there is no need to split a separate test set.
We exploit a high-quality Lanczos filter~\cite{Lanczos} for image resizing and save images in the uncompressed PNG format.
We will detail each dataset separately as follows.

\begin{itemize}
	\item \textbf{AFHQ-Cat.} We use the AFHQ-Cat dataset released by~\cite{starganv2}, which is the cat category of the high-quality Animal Faces-HQ (AFHQ) dataset~\cite{starganv2}. The original authors mentioned that they collected images with permissive licenses from the Flickr and Pixabay websites. All images were vertically and horizontally aligned at the center. The dataset was released under Creative Commons BY-NC 4.0 license by NAVER Corporation. As a small dataset, we exploit $5,153$ training images of cat faces with various breeds. The original resolution is $512 \times 512$, which is scaled to $256 \times 256$ for training.

	\item \textbf{FFHQ.} The full Flickr-Faces-HQ (FFHQ) dataset~\cite{stylegan} consists of $70,000$ high-quality ($1024 \times 1024$) human face images. The faces contain considerable variation in terms of age, ethnicity, background, and accessories. The original authors mentioned that the images were crawled from the Flickr website, and only the ones under permissive licenses were collected. The images were automatically aligned~\cite{dlib} and cropped. The individual images were published in Flickr by their respective authors under either Creative Commons BY 2.0, Creative Commons BY-NC 2.0, Public Domain Mark 1.0, Public Domain CC0 1.0, or U.S. Government Works license. The dataset itself was made available under Creative Commons BY-NC-SA 4.0 license by NVIDIA Corporation. We randomly select different subsets of FFHQ, \ie, FFHQ-1k ($1,000$ images, $\sim1.4\%$ data), FFHQ-5k ($5,000$ images, $\sim7\%$ data), FFHQ-7k ($7,000$ images, $10\%$ data), and FFHQ-70k ($70,000$ images, $100\%$ data) to perform GAN training given different data amounts. The original images are resized to $256 \times 256$ for training in the experiments of our main paper. We will also show some synthesized examples with the original resolution in this Appendix.

	\item \textbf{Danbooru2019 Portraits (Anime).} The Danbooru2019 Portraits (Anime)~\cite{danbooru2019portraits} is a dataset consisting of $512 \times 512$ anime faces cropped from solo SFW Danbooru2019 images~\cite{danbooru2019}. The full dataset contains a total of $302,652$ images in a broad portrait style encompassing ears, necklines, hats, \etc., rather than the tightly cropped faces. The dataset was released under the Creative Commons public domain (CC-0) license. We artificially confine the dataset into a subset for GAN training, \ie, Anime-5k ($5,000$ images, $\sim2\%$ data). The training image size is $256 \times 256$.

	\item \textbf{Caltech-UCSD Birds-200-2011 (CUB).} The Caltech-UCSD Birds-200-2011 (CUB)~\cite{cub2002011} is an extended version of CUB-200~\cite{cub200}, a challenging dataset containing $200$ bird species. The dataset consists of $11,788$ bird images at diverse locations with heavy background clutter. The images were harvested using the Flickr image search. We were unable to find the license information about this dataset. We employ all the available images for GAN training since the dataset is already small and difficult. The resolutions of original images vary, and we uniformly resize them to $256 \times 256$.

\end{itemize}

\section{Implementation Details}
\label{sec:implementation}
In the main paper, we choose the state-of-the-art StyleGAN2~\cite{stylegan2} as the backbone to verify the effectiveness of the proposed APA on limited data. Besides, we compare our method with representative approaches designed for the low-data regime, including the adaptive discriminator augmentation (ADA)~\cite{ada} and LC-regularization (LC-Reg)~\cite{reglimited}, which perform standard data augmentations and model regularization, respectively. In addition, we compare APA with representative conventional techniques for regularizing GANs, \ie, instance noise~\cite{noisestable2} and one-sided label smoothing~\cite{improvedtechgans}. For a fair and controllable comparison, we reimplement all baselines and run the experiments from scratch using official code. The qualitative and quantitative results of each method are reported using the best model throughout training.

We ran our experiments on an internal computing cluster with Slurm Workload Manager. All the models are trained on 8 NVIDIA Tesla V100 GPUs with 32 GB memory capacity. We follow~\cite{ada} and employ its mixed-precision FP16/FP32 training scheme in all our experiments. The actual memory consumption for each model is around 11 GB per GPU. We perform $25,000$ kimg (\ie, thousands of images shown to the discriminator, measuring the training progress~\cite{stylegan,stylegan2,ada}) of training for each model. For the average training time cost of different models, please refer to the Training cost of Section 4.2 in the main paper.

We implemented the proposed APA on top of the official implementation of StyleGAN2~\cite{stylegan2}. The network architecture is kept unchanged. The mapping network contains $8$ fully connected layers, and the dimensionality of the input and intermediate latent space is $512$. We use the combination of a generator with output skips and a residual discriminator. The detailed structures of the generator and the discriminator are the same as~\cite{stylegan2}, \eg, using weight demodulation~\cite{stylegan2} in the generator. The activation function is Leaky ReLU with a negative slope of $0.2$. We apply several other standard techniques in~\cite{pggan,stylegan,stylegan2}, including the mini-batch standard deviation layer at the end of the discriminator~\cite{pggan}, equalized learning rate for all the trainable parameters~\cite{pggan}, pixel-wise feature vector normalization~\cite{pggan}, the exponential moving average of generator weights~\cite{pggan}, style mixing regularization~\cite{stylegan}, path length regularization~\cite{stylegan2}, and lazy regularization~\cite{stylegan2}.
The training loss is the non-saturating logistic loss~\cite{GAN,stylegan2} with $R_1$ regularization~\cite{r1reg}. The batch size is $64$ for our experiments trained with the resolution of $256 \times 256$. The Adam~\cite{adam} optimizer is applied with $\beta_1=0, \beta_2=0.99$. For other network and training details, we follow the original paper and official code of StyleGAN2~\cite{stylegan2}.

For APA, we set the overfitting heuristic $\lambda=\lambda_{r}$ in our main experiments and study other variants (\ie, $\lambda=\lambda_{f}$ and $\lambda=\lambda_{rf}$) through the ablation study. Aside from the ablation study of a fixed deception probability $p=0.5$, $p$ is adaptively adjusted according to $\lambda$. The adaptive adjustment of $p$ is as follows. We first initialize $p$ to be zero and set a threshold value $t$ ($t=0.6$ in most cases unless specified otherwise). If $\lambda$ signifies too much/little overfitting regarding $t$ (\ie, larger/smaller than $t$), the probability $p$ will be increased/decreased by one fixed step. Using this step size, $p$ can increase quickly from zero to one, \ie, in $500$k images shown to $D$. We adjust $p$ once every four iterations. We clamp $p$ from below to zero after each adjustment so that $p$ can always be larger than zero. We do not set an upper bound for $p$ while it can be naturally restricted under a safe limit. Then, the pseudo augmentation of each instance will be applied with the probability $p$ or be skipped with the probability $1-p$. In this way, the strength of pseudo augmentation can be adaptively controlled based on the degree of overfitting throughout training.

As for other methods used for comparison, we strictly follow all the details in their papers and released official code using recommended setups.

\vspace{-0.05cm}
\section{More Results and Analysis}
\label{sec:examples}

\begin{table}[tb!]
\vspace{-0.26cm}
\centering
\small
\caption{The FID (lower is better) and IS (higher is better) scores ($256 \times 256$) of \textbf{transfer learning} on MetFaces~\cite{ada} with limited data amounts from the pre-trained StyleGAN2 model on FFHQ-70k~\cite{stylegan}.}
\begin{tabularx}{\textwidth}{l|*{4}{|Y}}
\Xhline{1pt}
& \multicolumn{2}{c}{MetFaces-1336 (full data)} & \multicolumn{2}{|c}{MetFaces-500 ($\sim37\%$ data)} \\
\cline{2-5}
Method& FID $\downarrow$& IS $\uparrow$& FID $\downarrow$& IS $\uparrow$ \\
\Xhline{0.6pt}
StyleGAN2~\cite{stylegan2} & 30.988& 3.719& 54.691& 3.218 \\
APA (Ours)& {\bf21.050}& {\bf4.103}& {\bf29.508}& {\bf3.986} \\
\Xhline{1pt}
\end{tabularx}
\label{tbl:translearn}
\vspace{-0.14cm}
\end{table}

\textbf{Effectiveness of APA for transfer learning.}
Table~\ref{tbl:translearn} reports the additional transfer learning results on MetFaces~\cite{ada} from the pre-trained StyleGAN2 model on FFHQ-70k~\cite{stylegan}.
All the metrics can be boosted by APA, further verifying its effectiveness for transfer learning besides training from scratch.

\begin{table}[tb!]
\centering
\small
\caption{The FID (lower is better) and IS (higher is better) scores ($256 \times 256$) \textbf{under a lower amount of training data} on FFHQ-500~\cite{stylegan} (a subset of $500$ images, $\sim0.7\%$ of full data).}
\begin{tabularx}{\textwidth}{l|*{2}{|Y}}
\Xhline{1pt}
& \multicolumn{2}{|c}{FFHQ-500 ($\sim0.7\%$ data)} \\
\cline{2-3}
Method& FID $\downarrow$& IS $\uparrow$ \\
\Xhline{0.6pt}
StyleGAN2~\cite{stylegan2} & 119.815& 2.446 \\
APA (Ours)& {\bf50.989}& {\bf4.099} \\
\Xhline{1pt}
\end{tabularx}
\label{tbl:fewerffhq}
\vspace{-0.55cm}
\end{table}

\textbf{The performance of APA under a lower amount of data (\ie, fewer than 1k).}
We have reported transfer learning results on MetFaces-500~\cite{ada} (a subset of $500$ images, $\sim37\%$ of full data) in Table~\ref{tbl:translearn}.
Table~\ref{tbl:fewerffhq} shows more results on FFHQ-500~\cite{stylegan} (a subset of $500$ images, $\sim0.7\%$ of full data). 
Under fewer data, APA can still boost StyleGAN2 performance by a large margin. However, the quality itself of synthesized images by APA can still be improved, consistent with our discussion on limitations in the main paper.

\begin{figure}
	\centering
	\includegraphics[width=\linewidth]{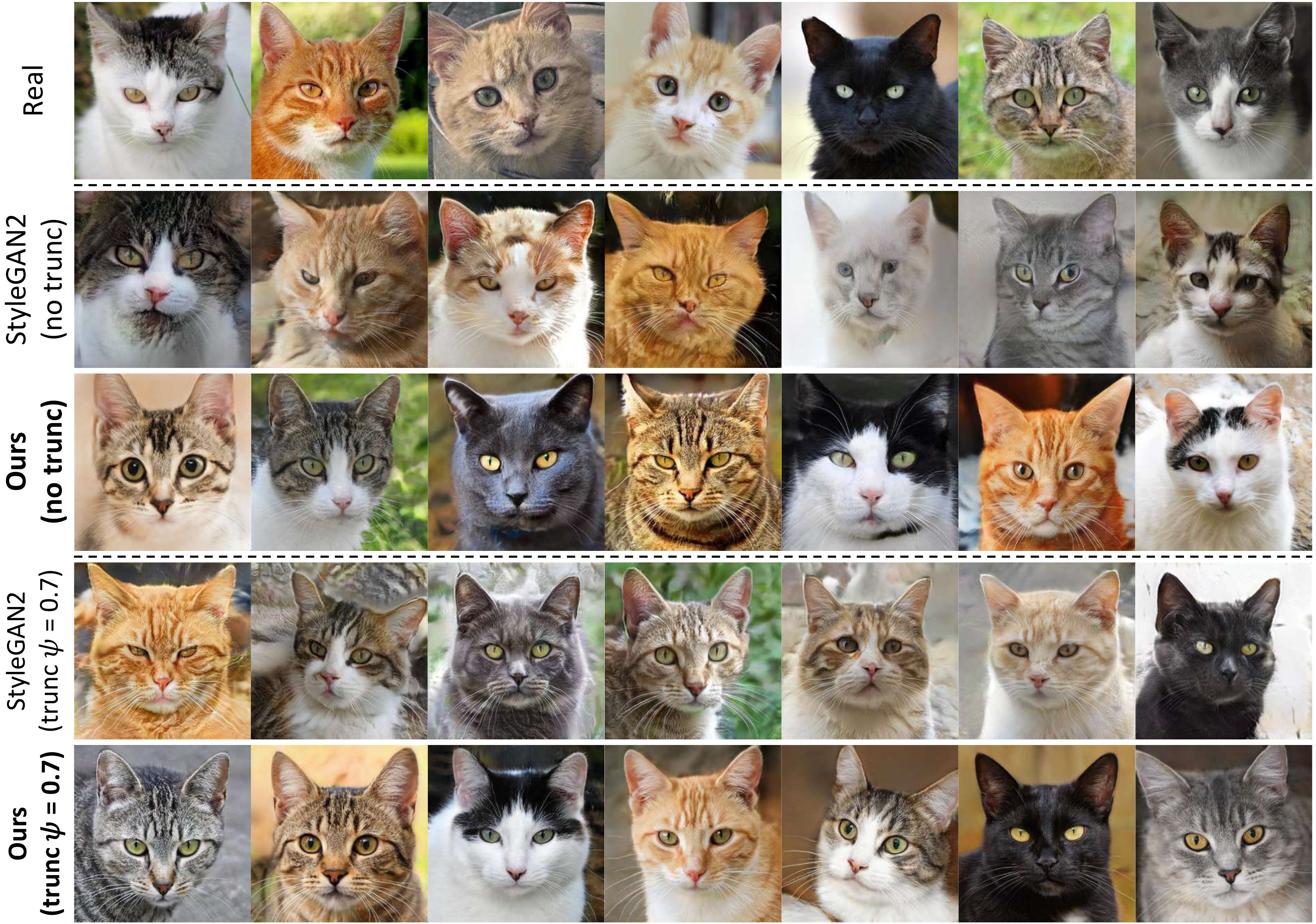}
	\caption{More examples of the effectiveness of our method to improve state-of-the-art StyleGAN2~\cite{stylegan2} synthesized results ($256 \times 256$) on \textbf{AFHQ-Cat-5k}~\cite{starganv2} ($5,153$ images, which is small by itself).}
	\label{fig:afhqmore}
\end{figure}

\begin{figure}
	\centering
	\includegraphics[width=\linewidth]{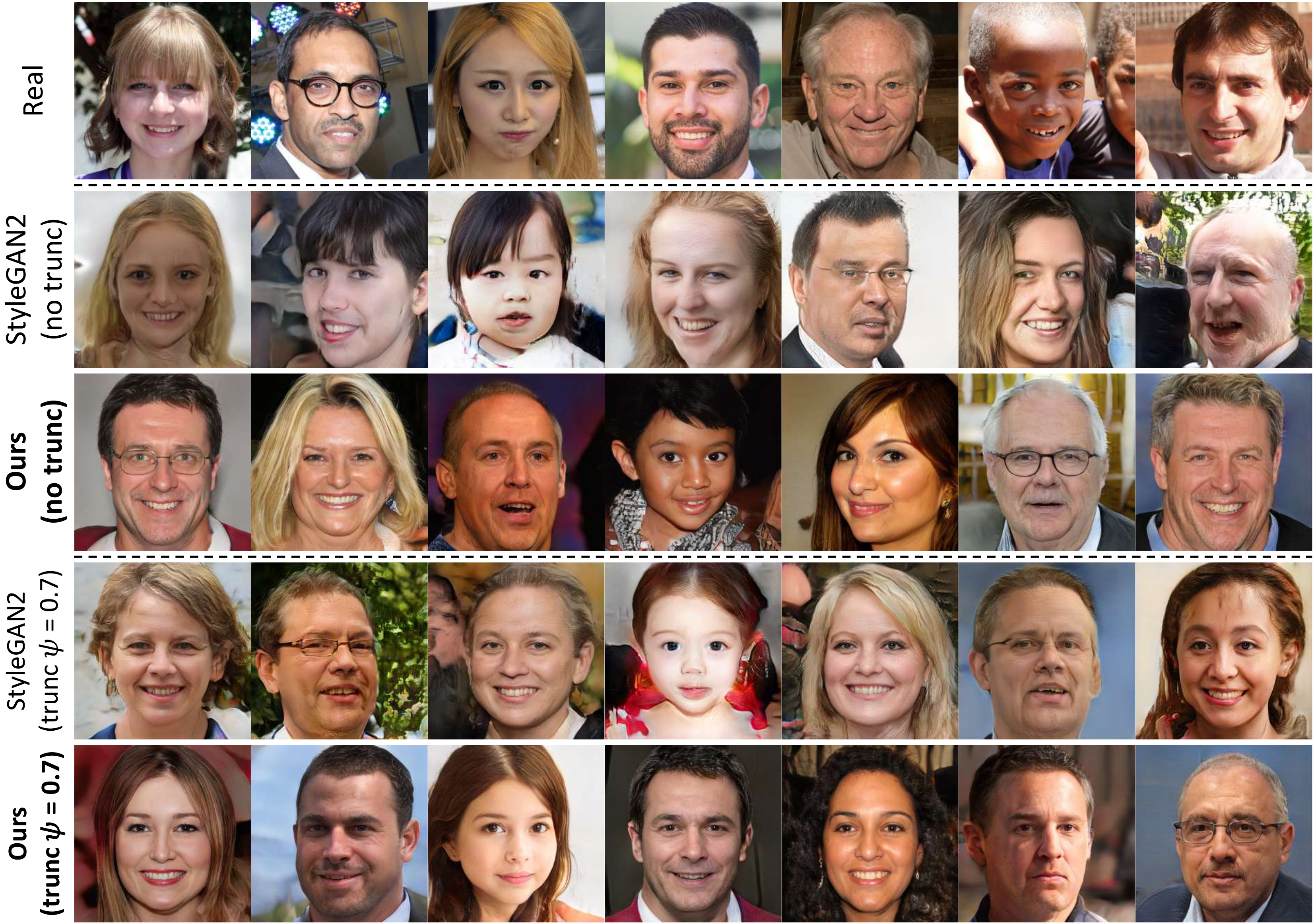}
	\caption{More examples of the effectiveness of our method to improve state-of-the-art StyleGAN2~\cite{stylegan2} synthesized results ($256 \times 256$) on \textbf{FFHQ-5k}~\cite{stylegan} (a subset of $5,000$ images, $\sim7\%$ of full data).}
	\label{fig:ffhqmore}
\end{figure}

\begin{figure}
	\centering
	\includegraphics[width=\linewidth]{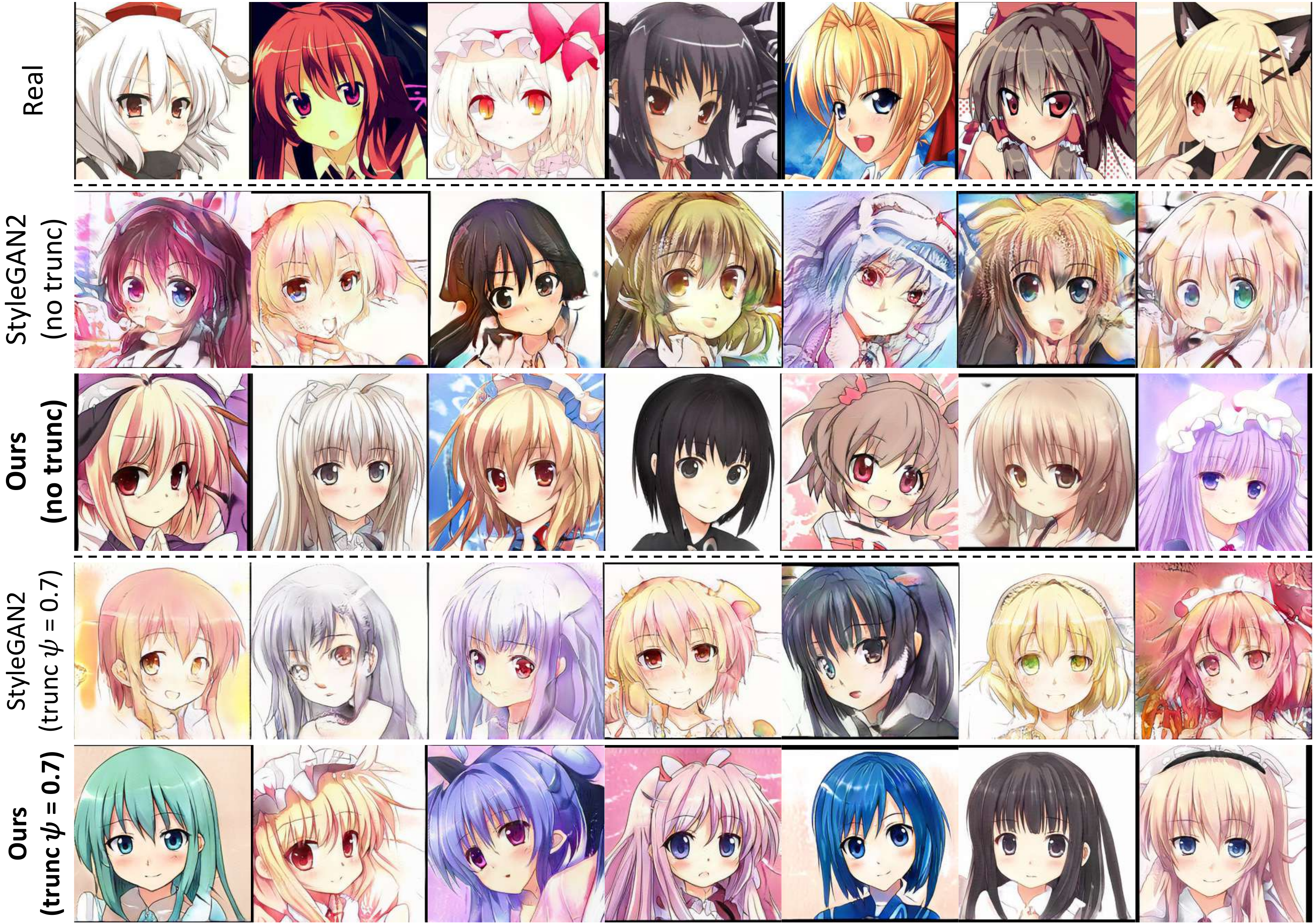}
	\caption{More examples of the effectiveness of our method to improve state-of-the-art StyleGAN2~\cite{stylegan2} synthesized results ($256 \times 256$) on \textbf{Anime-5k}~\cite{danbooru2019portraits} (a subset of $5,000$ images, $\sim2\%$ of full data).}
	\label{fig:animemore}
\end{figure}

\begin{figure}
	\centering
	\includegraphics[width=\linewidth]{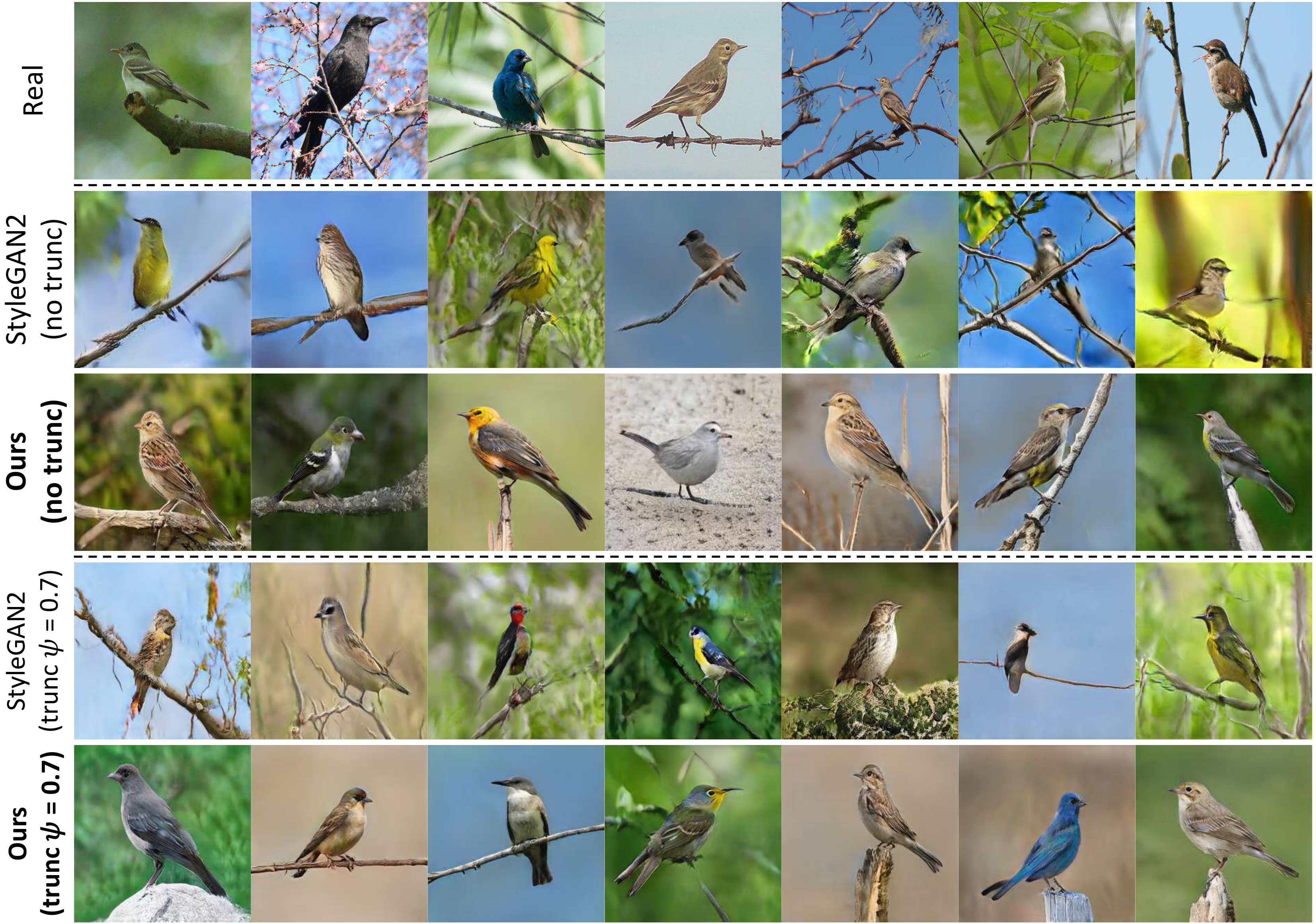}
	\caption{More examples of the effectiveness of our method to improve state-of-the-art StyleGAN2~\cite{stylegan2} synthesized results ($256 \times 256$) on \textbf{CUB-12k}~\cite{cub2002011} ($11,788$ images, which is small by itself).}
	\label{fig:cubmore}
\end{figure}

\textbf{More examples of the effectiveness of APA on various datasets.}
We show more comparative results of StyleGAN2~\cite{stylegan2} and the proposed APA to demonstrate the effectiveness of our method to improve the state-of-the-art baseline on various datasets with limited data amounts, \ie, AFHQ-Cat-5k~\cite{starganv2} (Figure~\ref{fig:afhqmore}), FFHQ-5k~\cite{stylegan} (Figure~\ref{fig:ffhqmore}), Anime-5k~\cite{danbooru2019portraits} (Figure~\ref{fig:animemore}), and CUB-12k~\cite{cub2002011} (Figure~\ref{fig:cubmore}).
Regardless of applying the truncation trick~\cite{BigGAN,stylegan,stylegan2} or not, the proposed APA can significantly ameliorate the degraded synthesis quality of StyleGAN2 with limited training data in all cases. The generated images by our method are highly photorealistic using only limited training data, being closer to the real data distributions.

\begin{figure}
	\centering
	\vspace{-0.23cm}
	\includegraphics[width=\linewidth]{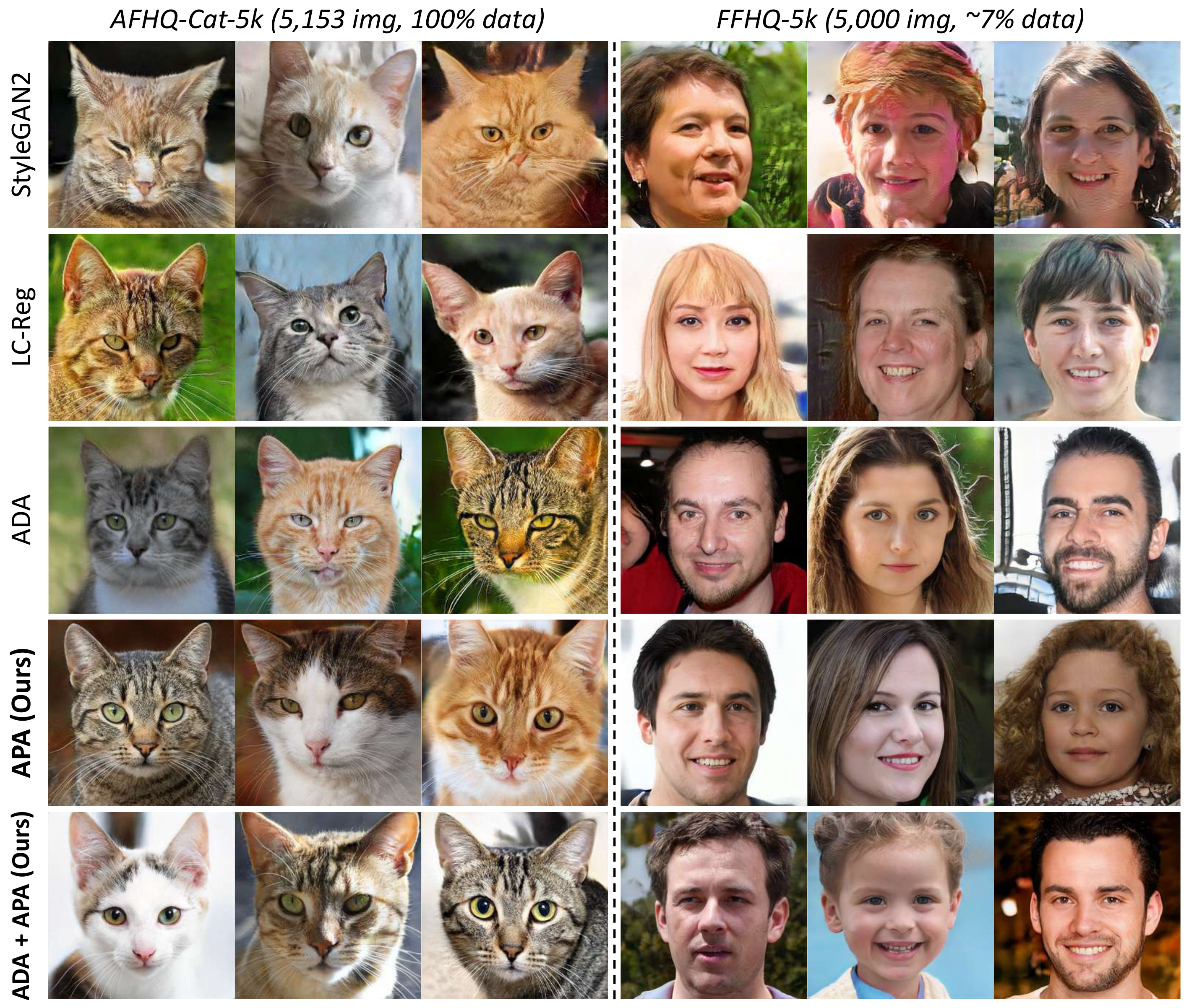}
	\vspace{-0.56cm}
	\caption{The synthesized results ($256 \times 256$, no truncation) of our method \textbf{compared to other state-of-the-art solutions designed for GAN training with limited data} on StyleGAN2~\cite{stylegan2}. We confine the data amount of FFHQ~\cite{stylegan} and directly use AFHQ-Cat~\cite{starganv2} that is small by itself.}
	\label{fig:compare}
	\vspace{-0.1cm}
\end{figure}

\textbf{Qualitative results of comparison with other solutions for GAN training with limited data.}
In the main paper, we have quantitatively shown the effectiveness of the proposed APA over other state-of-the-art approaches designed for the low-data regime, including ADA~\cite{ada} and LC-regularization (LC-Reg)~\cite{reglimited}, which perform standard data augmentations and model regularization, respectively. 
In Figure~\ref{fig:compare}, we show the qualitative results for further illustration. The synthesis quality of the StyleGAN2~\cite{stylegan2} baseline deteriorates on the limited amount of training data. Ripple artifacts and substantial distortions appear on the generated images by StyleGAN2. LC-Reg~\cite{reglimited} slightly improves the quality of synthesis by reducing the distortions on the images. The amelioration of visual quality is more evident by applying ADA~\cite{ada}, where the ripple artifacts are clearly subsided while some minor artifacts exist on the hair and beard. The proposed APA achieves comparable or even better visual quality than LC-Reg~\cite{reglimited} and ADA~\cite{ada}, effectively improving the StyleGAN2 synthesized results on limited data. Notably, APA is also complementary to ADA~\cite{ada} for gaining a further performance boost, suggesting the compatibility of our method with standard data augmentations.
The visual results we present here are in line with the quantitative performance in our main paper.

\begin{table}[tb!]
\centering
\small
\caption{The FID (lower is better) and IS (higher is better) scores ($256 \times 256$) of \textbf{additional comparison with the state-of-the-art ADA~\cite{ada}} tailored for GAN training with limited data on StyleGAN2 trained with FFHQ~\cite{stylegan} and MetFaces~\cite{ada}. The bold number indicates the best value, and the underline marks the second best.}
\begin{tabularx}{\textwidth}{l|*{8}{|Y}}
\Xhline{1pt}
& \multicolumn{2}{c}{FFHQ-7k} & \multicolumn{2}{|c}{FFHQ-1k} & \multicolumn{2}{|c}{MetFaces-1336} & \multicolumn{2}{|c}{MetFaces-500} \\
\cline{2-9}
Method& FID $\downarrow$& IS $\uparrow$& FID $\downarrow$& IS $\uparrow$& FID $\downarrow$& IS $\uparrow$& FID $\downarrow$& IS $\uparrow$ \\
\Xhline{0.6pt}
StyleGAN2~\cite{stylegan2} & 27.738& 4.264& 86.407& 2.806& 30.988& 3.719& 54.691& 3.218 \\
ADA~\cite{ada}& \underline{10.275}& 4.813& \underline{22.590}& \underline{4.239}& \underline{20.834}& 4.005& 30.368& 3.974 \\
\cline{1-9}
APA (Ours)& 10.800& \underline{4.860}& 45.192& 4.130& 21.050& \underline{4.103}& \underline{29.508}& \underline{3.986} \\
ADA + APA (Ours)& {\bf7.333}& {\bf4.994}& {\bf18.892}& {\bf4.316}& {\bf18.865}& {\bf4.207}& {\bf28.408}& {\bf4.044} \\
\Xhline{1pt}
\end{tabularx}
\label{tbl:compareada}
\vspace{-0.55cm}
\end{table}

\begin{figure}
	\centering
	\includegraphics[width=\linewidth]{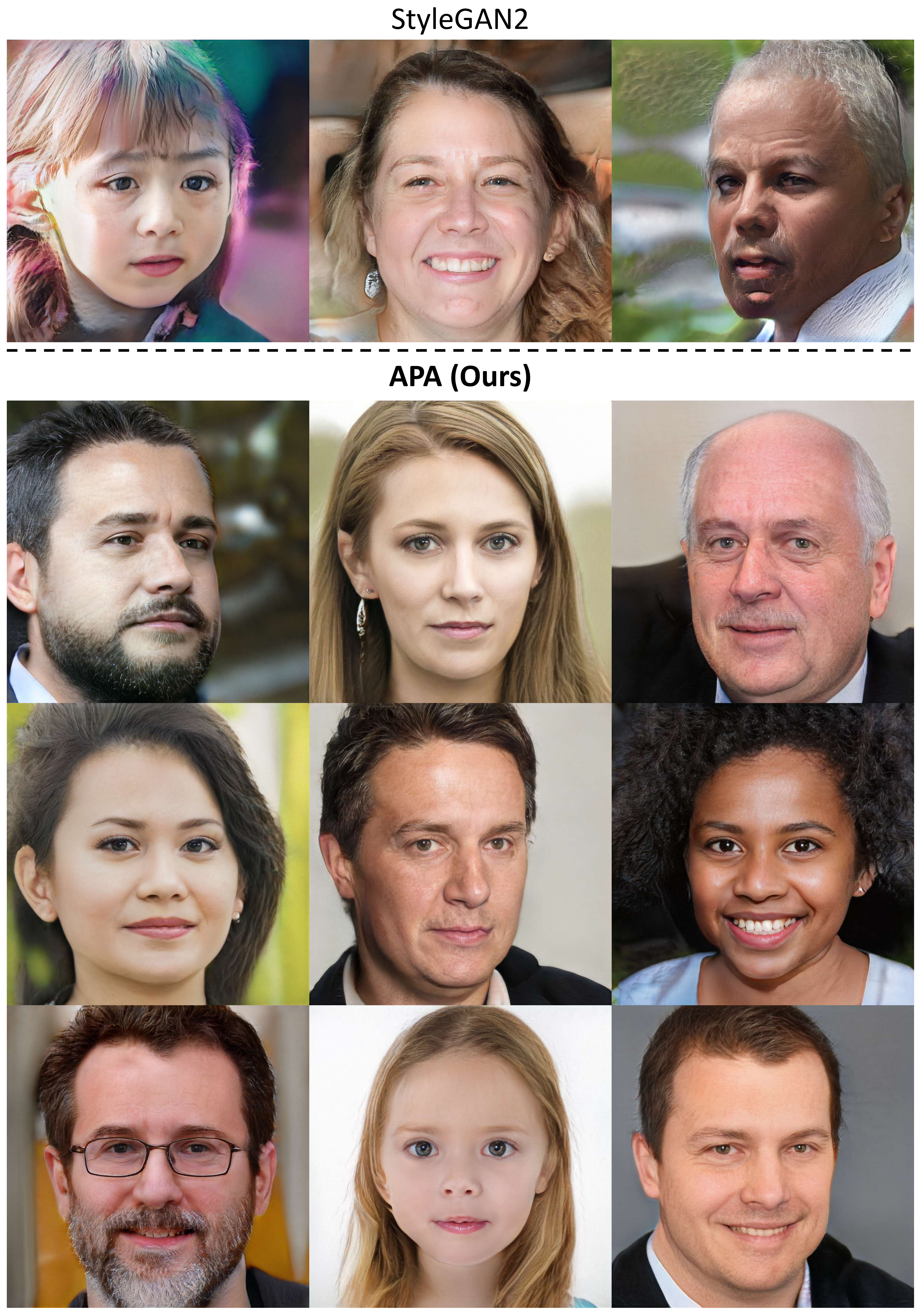}
	\caption{The effectiveness of our method to improve state-of-the-art \textbf{StyleGAN2}~\cite{stylegan2} higher-resolution synthesized results ($1024 \times 1024$, no truncation) \textbf{trained with the limited data}, \ie, FFHQ-5k~\cite{stylegan}~(a subset of $5,000$ images, $\sim7\%$ of full data). Our method achieves an FID score of $\bm{9.545}$, outperforming the original StyleGAN2 of $18.296$.}
	\label{fig:1024}
\end{figure}

\textbf{More comparative results with ADA~\cite{ada}.}
Table~\ref{tbl:compareada} reports additional comparative results with the state-of-the-art ADA~\cite{ada} tailored for GAN training with limited data to further highlight the advantages of the proposed APA.
We include additional comparisons on FFHQ-7k~\cite{stylegan} and FFHQ-1k (see our main paper for 5k and 70k). To make the comparison settings more comprehensive, we also provide the transfer learning results on MetFaces~\cite{ada} from the pre-trained StyleGAN2 model on FFHQ-70k. We also make a comparison with a fewer data amount (\ie, $500$ images).
Combined with the results we presented in the main paper, the proposed APA achieves comparable or even better performance than ADA~\cite{ada} while with less computational cost. Both methods outperform the StyleGAN2 baseline under limited data.
Although applying APA solely may be inferior to ADA~\cite{ada} on FFHQ-1k (in line with our discussion in the main paper), it is worth mentioning that APA is also complementary to ADA~\cite{ada}, which is very important to boost the performance further.

\textbf{Higher-resolution examples on StyleGAN2.}
In Figure~\ref{fig:1024}, we show some higher-resolution ($1024 \times 1024$) synthesized images on FFHQ-5k~\cite{stylegan}~(a subset of $5,000$ images, $\sim7\%$ of full data) to further illustrate the effectiveness of our approach in improving StyleGAN2 with limited training data. The truncation trick~\cite{BigGAN,stylegan,stylegan2} is not applied.
The proposed APA evidently improves StyleGAN2 in both perceptual quality and the FID score, indicating its effectiveness for GAN training with limited data.

\begin{figure}
	\centering
	\vspace{-0.2cm}
	\includegraphics[width=0.67\linewidth]{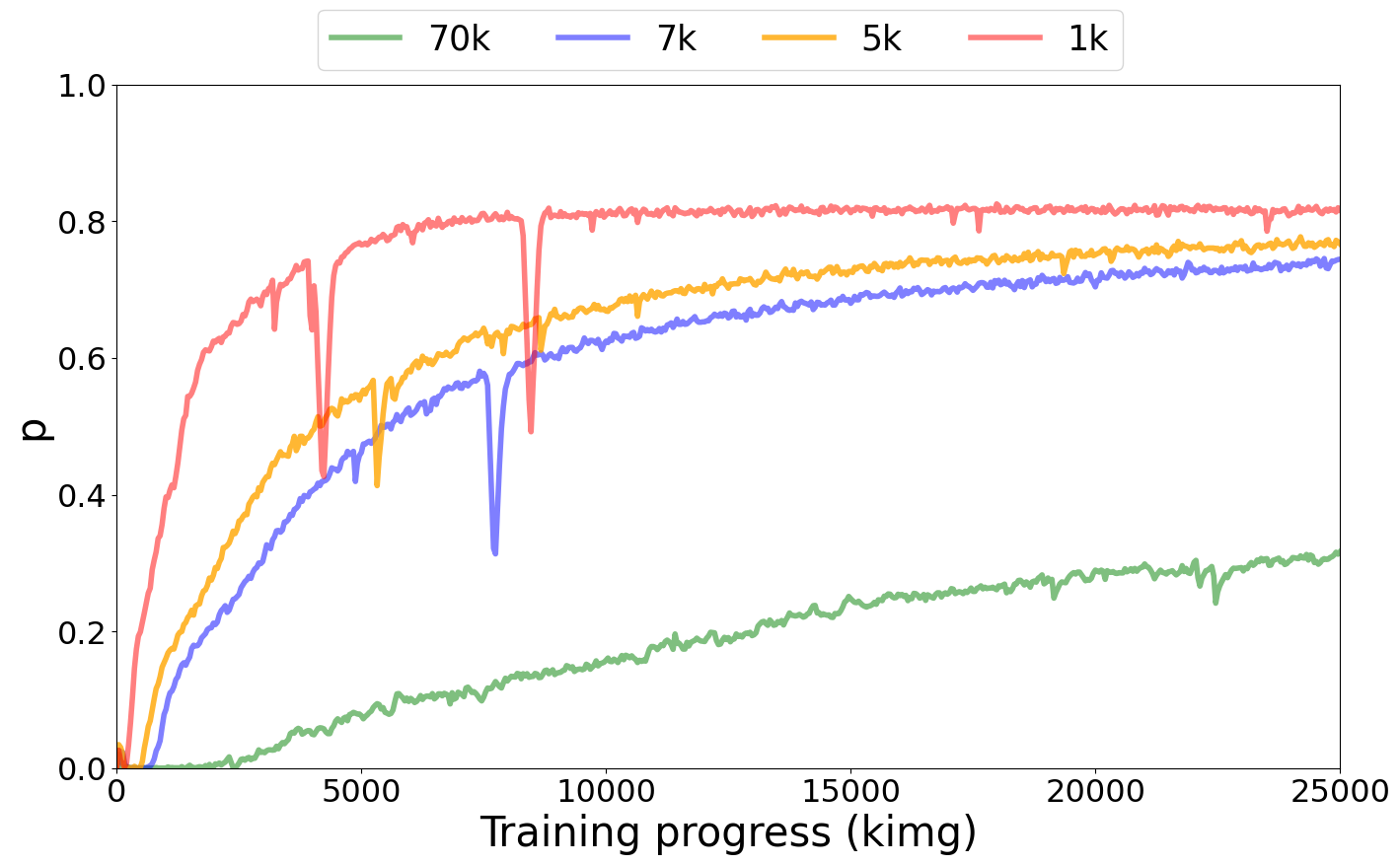}
	\vspace{-0.2cm}
	\caption{The \textbf{evolution of the deception probability $\bm p$} during training on FFHQ~\cite{stylegan} ($256 \times 256$) with different data amounts. The ``kimg'' denotes thousands of real images shown to the discriminator.}
	\label{fig:p}
	\vspace{-0.1cm}
\end{figure}

\textbf{Evolution of the deception probability.}
In Figure~\ref{fig:p}, we visualize the evolution of the deception probability $p$ in training on FFHQ~\cite{stylegan} ($256 \times 256$) with different data amounts. The evolution of $p$ may be relatively more unstable when the data amount is very limited (\ie, 1k). Notably, the proposed APA possesses a desired property that the deception probability $p$ can be naturally restricted under a safe limit~($\sim0.8$) regardless of the training data amounts. Hence, the fundamental capability of discriminator in adversarial training may be better preserved thanks to the adaptive control scheme.

\begin{figure}
	\centering
	\includegraphics[width=\linewidth]{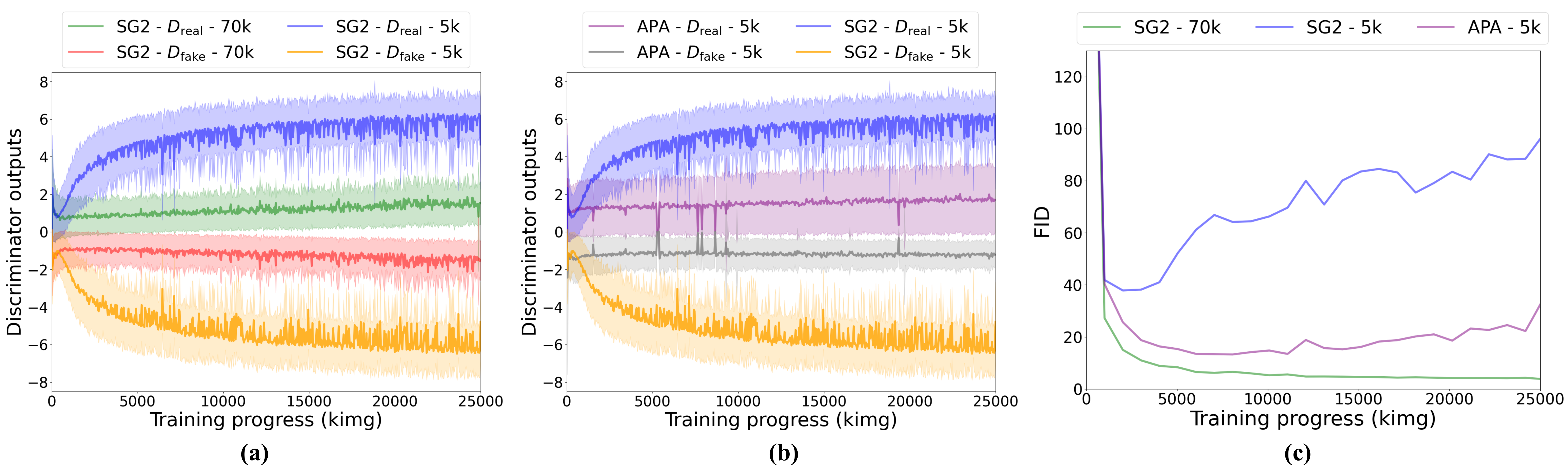}
	\vspace{-0.55cm}
	\caption{The \textbf{overfitting and convergence status} of APA compared to StyleGAN2 (SG2) on FFHQ~\cite{stylegan} ($256 \times 256$). (a) The discriminator raw output logits of StyleGAN2 on the full (70k) or limited (\textbf{5k}) datasets. (b) The discriminator raw output logits of StyleGAN2 and APA on the limited (\textbf{5k}) dataset. (c) The training convergence shown by FID.}
	\label{fig:overfitsg2apa5k}
	\vspace{-0.5cm}
\end{figure}

\begin{figure}
	\centering
	\vspace{-0.2cm}
	\includegraphics[width=\linewidth]{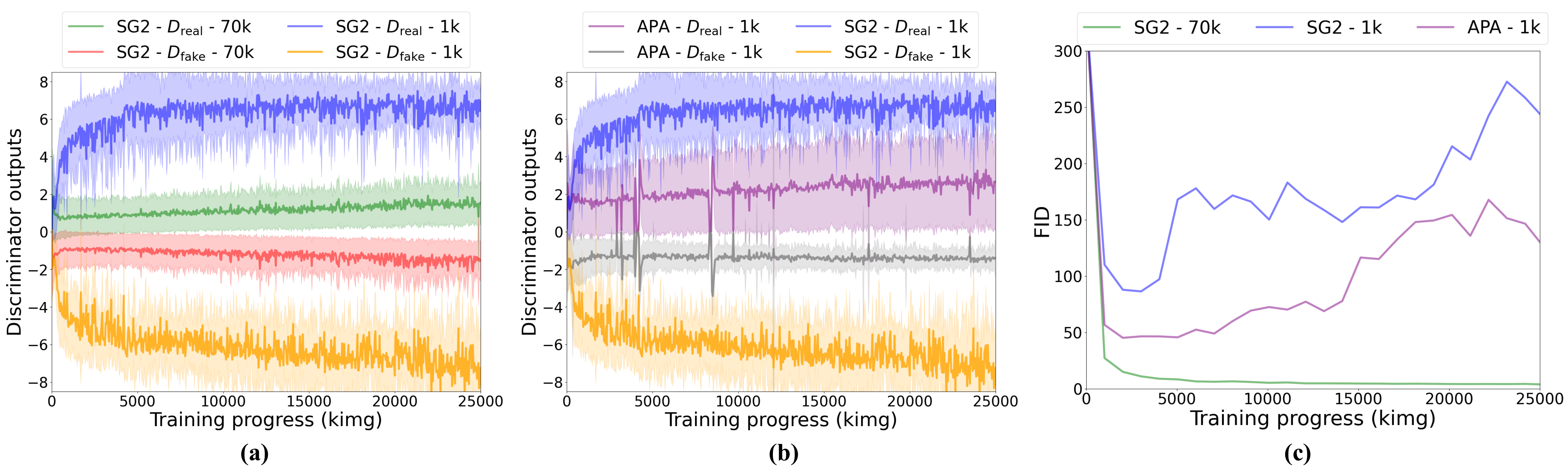}
	\vspace{-0.5cm}
	\caption{The \textbf{overfitting and convergence status} of APA compared to StyleGAN2 (SG2) on FFHQ~\cite{stylegan} ($256 \times 256$). (a) The discriminator raw output logits of StyleGAN2 on the full (70k) or limited (\textbf{1k}) datasets. (b) The discriminator raw output logits of StyleGAN2 and APA on the limited (\textbf{1k}) dataset. (c) The training convergence shown by FID.}
	\label{fig:overfitsg2apa1k}
	\vspace{-0.1cm}
\end{figure}

\begin{figure}
	\centering
	\includegraphics[width=\linewidth]{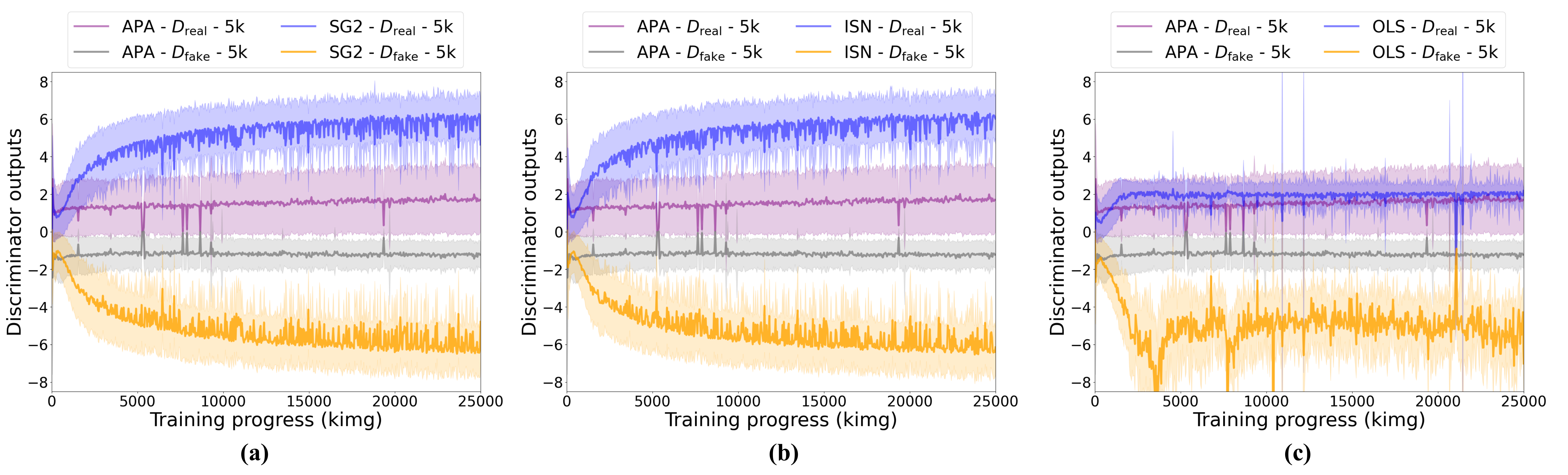}
	\vspace{-0.5cm}
	\caption{The \textbf{overfitting and convergence status} of APA compared to previous techniques for regularizing GANs on FFHQ-5k~\cite{stylegan} ($256 \times 256$, $\sim7\%$ data). (a) The discriminator raw output logits of StyleGAN2 (SG2) and APA. (b) The discriminator raw output logits of instance noise (ISN)~\cite{noisestable2} and APA. (c) The discriminator raw output logits of one-sided label smoothing (OLS)~\cite{improvedtechgans} and APA.}
	\label{fig:overfittricksapa}
	\vspace{-0.1cm}
\end{figure}

\begin{table}[tb!]
\centering
\small
\caption{The FID (lower is better) and IS (higher is better) scores ($32 \times 32$) of \textbf{class-conditional image synthesis on BigGAN} trained with the subsets of \textbf{CIFAR-10}~\cite{cifar} with limited data amounts.}
\begin{tabularx}{\textwidth}{l|*{6}{|Y}}
\Xhline{1pt}
& \multicolumn{2}{c}{full data} & \multicolumn{2}{|c}{$20\%$ data} & \multicolumn{2}{|c}{$10\%$ data} \\
\cline{2-7}
Method& FID $\downarrow$& IS $\uparrow$& FID $\downarrow$& IS $\uparrow$& FID $\downarrow$& IS $\uparrow$ \\
\Xhline{0.6pt}
BigGAN~\cite{BigGAN} & 9.531& 9.078& 22.024& 8.343& 44.061& 7.589 \\
APA (Ours)& {\bf8.283}& {\bf9.362}& {\bf15.316}& {\bf8.822}& {\bf25.987} & {\bf	8.410} \\
\Xhline{1pt}
\end{tabularx}
\label{tbl:biggancifar10}
\vspace{-0.1cm}
\end{table}

\begin{table}[tb!]
\centering
\small
\caption{The FID (lower is better) and IS (higher is better) scores ($32 \times 32$) of \textbf{class-conditional image synthesis on BigGAN} trained with the subsets of \textbf{CIFAR-100}~\cite{cifar} with limited data amounts.}
\begin{tabularx}{\textwidth}{l|*{6}{|Y}}
\Xhline{1pt}
& \multicolumn{2}{c}{full data} & \multicolumn{2}{|c}{$20\%$ data} & \multicolumn{2}{|c}{$10\%$ data} \\
\cline{2-7}
Method& FID $\downarrow$& IS $\uparrow$& FID $\downarrow$& IS $\uparrow$& FID $\downarrow$& IS $\uparrow$ \\
\Xhline{0.6pt}
BigGAN~\cite{BigGAN} & 13.281& 10.525& 35.590& 8.706& 64.828& 6.635 \\
APA (Ours)& {\bf11.429}& {\bf11.243}& {\bf23.506}& {\bf9.811}& {\bf45.794} & {\bf	8.114} \\
\Xhline{1pt}
\end{tabularx}
\label{tbl:biggancifar100}
\vspace{-0.55cm}
\end{table}

\textbf{More overfitting and convergence analysis.}
Aside from the overfitting and convergence analysis on FFHQ-7k ($10\%$ of full data) in our main paper, we present the additional analysis on FFHQ-5k ($\sim7\%$ of full data) and FFHQ-1k ($\sim1.4\%$ of full data) in Figure~\ref{fig:overfitsg2apa5k} and Figure~\ref{fig:overfitsg2apa1k}, respectively.
The trend of model overfitting and convergence on FFHQ-5k and FFHQ-1k is similar to FFHQ-7k. The divergence of StyleGAN2 discriminator predictions can be effectively restricted by applying the proposed APA, indicating its effectiveness in mitigating the discriminator overfitting. Meanwhile, APA improves the training convergence measured by FID.

Furthermore, we show the comparative overfitting analysis on APA and previous techniques for regularizing GANs in Figure~\ref{fig:overfittricksapa}.
We observe that the effectiveness to counteract overfitting is in line with the generation performance of these methods in the main paper.
Specifically, StyleGAN2 experiences diverged predictions most rapidly, and APA obtains the most effective restriction on the divergence of discriminator outputs. Applying instance noise (ISN)~\cite{noisestable2} produces curves that are very close to StyleGAN2, and one-sided label smoothing (OLS)~\cite{improvedtechgans} only restricts the real predictions.
This further suggests: 1) the importance of addressing the discriminator overfitting for training GANs with limited data; 2) the effectiveness of APA in alleviating the discriminator overfitting, outperforming previous strategies.

\textbf{Additional training convergence visualizations.}
Please refer to our \href{https://liming-jiang.com/projects/APA/resources/suppvideo.mp4}{\color{magenta}supplementary video} for additional training convergence visualizations on FFHQ-7k~\cite{stylegan} ($7,000$ images, $10\%$ of full data). The truncation trick~\cite{BigGAN,stylegan,stylegan2} with $\psi=0.7$ is applied to the synthesized images. The proposed APA effectively improves the training convergence of StyleGAN2 on limited data.

\textbf{Class-conditional image synthesis with BigGAN~\cite{BigGAN}.}
To further enrich our benchmark and demonstrate the effectiveness of the proposed APA under diverse settings, we show additional class-conditional image synthesis results on the state-of-the-art BigGAN~\cite{BigGAN} trained with CIFAR-10~\cite{cifar} and CIFAR-100~\cite{cifar} in Table~\ref{tbl:biggancifar10} and Table~\ref{tbl:biggancifar100}, respectively.
It can be observed that APA outperforms BigGAN under limited training data in all cases, further suggesting its adaptability and effectiveness for class-conditional image synthesis with other powerful contemporary GANs, such as BigGAN.

\end{document}